\documentclass{article}

\usepackage{arxiv}
\usepackage{natbib}
\usepackage{bm}

\usepackage[utf8]{inputenc}
\usepackage[T1]{fontenc}
\usepackage{hyperref}
\usepackage{url}
\usepackage{booktabs}
\usepackage{amsfonts}
\usepackage{nicefrac}
\usepackage{microtype}
\usepackage{xcolor}

\usepackage{graphicx}
\usepackage{xspace}
\usepackage{mathtools}
\usepackage{amsmath}
\usepackage{amsthm}
\usepackage{amsfonts}
\usepackage{caption}
\usepackage{subcaption}
\usepackage{multirow}

\usepackage{algorithmic}
\usepackage{algorithm}
\usepackage{booktabs}
\usepackage{float}
\newtheorem{lemma}{Lemma}

\newtheorem{theorem}{Theorem}
\newtheorem{proposition}{Proposition}
\newtheorem{remark}{Remark}
\newtheorem{definition}{Definition}

\title{On Reward Transferability in Adversarial Inverse Reinforcement Learning: Insights from Random Matrix Theory\thanks{This work was supported in part by the National Natural Science Foundation of China under Grant 12301351. }}

\author{
Yangchun Zhang$^{1}$, Wang Zhou$^{2}$, Yirui Zhou$^{1}$ \\
$^1$Department of Mathematics, College of Sciences, Shanghai University \\
$^2$Department of Statistics and Data Science, National University of Singapore
}

\begin{document}
\maketitle

\begin{abstract}
In the context of inverse reinforcement learning (IRL) with a single expert, adversarial inverse reinforcement learning (AIRL) serves as a foundational approach to providing comprehensive and transferable task descriptions. However, AIRL faces practical performance challenges, primarily stemming from the framework's overly idealized decomposability condition, the unclear proof regarding the potential equilibrium in reward recovery, or questionable robustness in high-dimensional environments. This paper revisits AIRL in \textbf{high-dimensional scenarios where the state space tends to infinity}. Specifically, we first establish a necessary and sufficient condition for reward transferability by examining the rank of the matrix derived from subtracting the identity matrix from the transition matrix. Furthermore, leveraging random matrix theory, we analyze the spectral distribution of this matrix, demonstrating that our rank criterion holds with high probability even when the transition matrices are unobservable. This suggests that the limitations on transfer are not inherent to the AIRL framework itself, but are instead related to the training variance of the reinforcement learning algorithms employed within it. Based on this insight, we propose a hybrid framework that integrates on-policy proximal policy optimization in the source environment with off-policy soft actor-critic in the target environment, leading to significant improvements in reward transfer effectiveness.
\end{abstract}

\noindent
{\it Keywords:} Adversarial inverse reinforcement learning, random matrix theory, spectral distribution, reward transfer, transferability condition

\section{Introduction\label{sec_intro}}
Imitation learning (IL) \citep{pomerleau1991efficient,ng2000algorithms,syed2007a,ho2016generative} efficiently trains a policy from expert demonstrations. Serving as a potent and practical alternative to reinforcement learning (RL) \citep{puterman2014markov,sutton2018reinforcement}, it eliminates the need for designing reward signals. IL has demonstrated significant success in diverse and complex domains, such as autonomous driving \citep{bhattacharyya2018multi}, robot manipulation \citep{jabri2021robot} and commodity search \citep{shi2019virtual}. 

Inverse reinforcement learning (IRL) \citep{ng2000algorithms}, within the field of IL, excels at handling the transfer paradigm by providing a more adaptable representation of the expert's task for policy training in environments with varying dynamics. However, the fact that rewards induced by a given optimal policy are not unique inhibits the recovery of the expert's true underlying reward \citep{ng1999policy}. A common approach to addressing this issue assumes the expert to be optimal under entropy-regularized RL \citep{ziebart2010modeling}, where the reward can be identified up to potential shaping transformations \citep{ng1999policy,cao2021identifiability}. While shaped rewards can increase learning speed in the original training environment, when the reward is deployed at test-time on environments with varying dynamics, it may no longer produce optimal behavior \citep{fu2018learning}. For better transferability across different dynamics, the reward should be identified up to a constant. \cite{fu2018learning} introduced the adversarial inverse reinforcement learning (AIRL) framework, ensuring reward identifiability up to a constant and facilitating effective transfer in dynamic changes through learning disentangled (state-only) rewards.

Nevertheless, recent IRL studies \citep{arnob2020off,xu2022receding,hoshino2022opirl} implementing AIRL have faced poor transfer performance in certain contexts. They attribute this inefficiency to the framework's overly idealized decomposability condition (functions over current states and next states can be isolated from their mixture) or to the unclear proof regarding the potential equilibrium in reward recovery (which is proved to be identified up to a constant). For more details, please refer to \citep{liu2020state,cao2021identifiability,liu2021energy}. Moreover, a practical challenge faced by AIRL (more broadly all IRL methods) is the high complexity of real-world environments, which means AIRL struggles to handle high-dimensional state spaces. Specifically, it is questioned whether AIRL remains robust when dealing with such cases.

This paper revisits reward transferability in high-dimensional frameworks \citep{BSbook}, where the state space size $|\mathcal{S}| \to \infty$, a common feature of complex real-world environments. From a rank perspective, we show that the reward can be identified up to a constant if ${\rm rank}(\mathbf{P} - I) = |\mathcal{S}| - 1$, which is both necessary and sufficient. Here, $\mathbf{P}$ denotes the transition matrix:
\begin{align}
\left[
\begin{array}{ccc}
p(s^{1}|s^{1}) & \ldots & p(s^{|\mathcal{S}|}|s^{1}) \\
\ldots & \ldots & \ldots \\
p(s^{1}|s^{|\mathcal{S}|}) & \ldots & p(s^{|\mathcal{S}|}|s^{|\mathcal{S}|})
\end{array}
\right], 
\label{transition kernel}
\end{align}
$p(s'|s)=\mathbb{E}_{a\sim \pi_{p}^{\star}}\left[ p(s'|s,a) \right]=\sum_{a}\pi_{p}^{\star}(a|s)p(s'|s,a)$, $\pi_{p}^{\star}$ is the optimal policy under the ground truth reward $r_{gt}$ and $p$, $I$ is the identity matrix. To prevent confusion with the time step specified in the subscript, here we employ superscripts to denote the state index. This raises a critical question: \textbf{Is the rank criterion robustly satisfied in high-dimensional state spaces}? (As the size of $|\mathcal{S}|$ grows to infinity, does exactly one singular value of the matrix $\mathbf{P}-I$ that tends to 0?) This question is further complicated by unobservable transition matrices, making it impractical to rely on a fixed transition matrix for theoretical analysis.

For an unobservable transition matrix $\mathbf{P}$, we adopt a variational inference approach (Bayesian statistics) \citep{efron2016computer,levine2018reinforcement} and assume the transition matrix $\mathbf{P}$ follows a probability model $\mathfrak{P}=\left(p_{i j}\right) \in \mathbb{R}^{|\mathcal{S}| \times|\mathcal{S}|}$ governed by a flat Dirichlet prior in the absence of prior information. We then examine the condition from a random matrix theory (RMT) perspective \citep{BSbook,Bao_2015}, and extend this analysis to cases with limited prior information. The RMT analysis reveals that $\mathfrak{P}-I$ has exactly one singular value equal to 0, while the remaining $|\mathcal{S}|-1$ singular values are at a distance of order at most $|\mathcal{S}|^{-1/4}$ from 1 (thus far from 0). Consequently, ${\rm rank}(\mathfrak{P}-I)=|\mathcal{S}|-1$ with high probability, ensuring that the reward is identified up to a constant and remains robust in high dimensions (since $|\mathcal{S}|-1$ singular values are not close to 0). Furthermore, we extend our analysis to cases where limited prior information is available, such as the locations of some obstacles, and obtain results similar to those under uninformative priors.

Based on the above robustness analysis, we shift our focus from the AIRL framework (the performance issues are not due to its design) to the RL algorithm employed during the training process. This change in focus leads to some surprising findings. We assert that the RL algorithm employed by AIRL is crucial for balancing recovery stability and training efficiency by quantifying the training variance. Our proposed solution involves distinctly analyzing on-policy and off-policy RL algorithms in the source and target environments. In the source environment, on-policy methods ensure stable training and accurate reward recovery by interacting directly with the current policy, avoiding distributional shifts that off-policy methods might introduce. Conversely, in the target environment, off-policy methods are more effective due to their sample efficiency, utilizing past experiences to enhance policy training, while on-policy methods may lead to inefficiency because of limited sample reuse. Recognizing that mismatches such as employing an off-policy algorithm in the source environment or an on-policy algorithm in the target environment are the primary sources of inefficiency, we propose a hybrid framework. This framework employs on-policy proximal policy optimization (PPO) \citep{schulman2017proximal} as the RL algorithm in the source environment with off-policy soft actor-critic (SAC) \citep{haarnoja2018soft2} in the target environment, referred to as PPO-AIRL + SAC, to significantly improve reward transfer effectiveness. 

The remainder of this paper is organized as follows. Section \ref{sec_related} briefly reviews the related work on robust reward recovery. In Section \ref{sec_back}, we provide the necessary background. Section \ref{sec_disentangled_condition_environment_dynamics} explores environments capable of extracting disentangled rewards. Next, we examine the extractability of these rewards under different policy optimization methods and propose the hybrid PPO-AIRL + SAC framework in Section \ref{sec_reward_transferable_analysis}. To validate the findings in Section \ref{sec_disentangled_condition_environment_dynamics} and the performance of PPO-AIRL + SAC, we conduct comprehensive experiments in Section \ref{sec_experiments}. Finally, Section \ref{sec_conc} concludes the paper.

\section{Related Work\label{sec_related}}
RL \citep{puterman2014markov,sutton2018reinforcement} employs the optimization of a cumulative future reward to develop policies that effectively address sequential decision problems. Unlike RL, which often requires the manual design of a reward function, IL \citep{pomerleau1991efficient,ng2000algorithms,syed2007a,ho2016generative} leverages expert demonstrations to train the agent. Within IL, IRL \citep{ng2000algorithms} has emerged as a significant approach. IRL, which focuses on inferring a reward function that leads to an optimal policy aligned with expert behavior, has shown remarkable effectiveness and attracted much attention in recent years \citep{fu2018learning,kostrikov2019discriminator,kostrikov2020imitation,zhou2023distributional}. A key challenge in IRL is recovering robust reward functions that can be effectively transferred to environments with varying dynamics \citep{fu2018learning,xu2022receding,zhou2024generalization}.

AIRL \citep{fu2018learning} provides for simultaneous learning of the reward and value function, which is robust to changes in dynamics. Empowerment-regularized adversarial inverse reinforcement learning (EAIRL) \citep{qureshi2019adversarial} learns empowerment by variational information maximization \citep{mohamed2015variational} to regularize the MaxEnt IRL, and subsequently obtains near-optimal rewards and policies. Their experimentation shows that the learned rewards are transferable to environments that are structurally or dynamically different from training environments. Drawing upon the analytic gradient of the $f$-divergence between the agent's and expert's state distributions (regarding reward parameters), \citep{ni2020f} devised $f$-IRL for retrieving a stationary reward function from the expert density via gradient descent. Based on the idea of distribution matching and AIRL, \cite{hoshino2022opirl} formulated off-policy inverse reinforcement learning (OPIRL), which not only improves sample efficiency but is able to generalize to unseen environments. Further, receding-horizon inverse reinforcement learning (RHIRL) \citep{xu2022receding} shows its superiority in scalability and robustness for high-dimensional, noisy, continuous systems with black-box dynamic models. It trains a state-dependent cost function ``disentangled'' from system dynamics under mild conditions to be robust against noise in expert demonstrations and system control. Unlike the MaxEnt framework that aims to maximize rewards around demonstrations, behavioral cloning inverse reinforcement learning (BC-IRL) \citep{szot2023bc} optimizes the reward parameter such that its trained policy matches the expert demonstrations better. 

Unlike these theoretically significant works, we concentrate on the practical scenario where the state transition matrix is unobservable in the absence of prior information (we also consider cases with limited prior knowledge). In this context, we demonstrate that the reward can be disentangled from an RMT perspective. Additionally, we argue that the RL algorithm employed by AIRL is essential for balancing recovery stability and training efficiency by quantifying training variance, a factor overlooked in previous research. In light of this, we propose a hybrid framework that optimizes solutions in the source environment while concurrently developing effective training strategies for new environments.

\section{Backgrounds and Notation\label{sec_back}}
\subsection{Markov Decision Process\label{sec_MDP}}
The interactions between the agent and the environment can be represented by the MDP \citep{puterman2014markov,sutton2018reinforcement} $(\mathcal{S}, \mathcal{A}, p, r_{gt}, \gamma, d_{0})$, with the state space $\mathcal{S}$, the action space $\mathcal{A}$, the transition dynamics $p(s'|s,a)$, the reward function $r(s,a)$, the discount factor $\gamma$ and the initial state distribution $d_{0}(s)$. The stochastic policy $\pi(a|s)$ is a probability function that maps a state $s \in \mathcal{S}$ to a distribution over actions $a \in \mathcal{A}$. The discounted stationary state distribution $d_{\pi}$ quantifies how often state $s$ is visited under policy $\pi$, which is defined as 
\begin{align*}
d_{\pi}(s)=(1-\gamma)\sum_{t=0}^{\infty}{\gamma ^{t}\mathbb{P}(s_{t}=s;\pi)}. 
\end{align*}
Analogously, the discounted stationary state-action distribution $\rho_{\pi}$ quantifies how often state-action pair $(s,a)$ is visited under policy $\pi$, and is defined as 
\begin{align*}
\rho_{\pi}(s,a)=(1-\gamma)\sum_{t=0}^{\infty}{\gamma ^{t}\mathbb{P}(s_{t}=s,a_{t}=a;\pi)}. 
\end{align*}

Denote $\pi^{\star}$ be the optimal policy that maximizes the entropy-regularized RL objective under $r$ and $p$:
\begin{align*}
\pi^{\star}=\mathop{\arg\max}_{\pi}\mathbb{E}_{\pi}\left[ \sum_{t=0}^{\infty}\gamma^{t}r(s_{t},a_{t}) + \alpha \mathbb{H}(\pi(\cdot|s_{t})) \right], 
\end{align*}
where $\mathbb{H}(\pi(\cdot|s_{t}))$ is the entropy of $\pi$ and $\alpha$ is the entropy temperature parameter. It can be demonstrated that the trajectory distribution generated by the optimal policy $\pi^{\star}(a|s)$ follows the form $\pi^{\star}(a|s) \propto \exp\{ Q_{r,p}^{\star}(s,a) \}$ \citep{ziebart2010modeling,haarnoja2017reinforcement}, where 
\begin{align*}
Q_{r,p}^{\star}(s,a)=\mathbb{E}_{\pi}\left[ \sum_{t'=t}^{\infty}\gamma^{t'}r(s_{t'},a_{t'}) + \alpha \mathbb{H}(\pi(\cdot|s_{t'})) | s_{t}=s,a_{t}=a \right]
\end{align*}
denotes the Q-function (soft). \\

\noindent \textbf{On-policy and off-policy RL policy optimization methods.} RL policy optimization methods typically update the Q-function and the policy $\pi$ iteratively, using samples collected by a behavior policy $\pi_{b}$. In on-policy RL algorithms, the behavior policy $\pi_{b}$ is the same as $\pi$, meaning the agent collects data from the environment using the same policy it is optimizing. In contrast, off-policy RL algorithms use the behavior policy $\pi_{b}$ to gather samples, while optimizing $\pi$ during training.

One of the key challenges in off-policy RL is accounting for the difference between $\pi_{b}$ and $\pi$ when updating the Q-function. To address this, off-policy RL algorithms introduce importance sampling. The importance sampling ratio 
\begin{align}
\rho_{t}=\frac{\pi(a_{t}|s_{t})}{\pi_{b}(a_{t}|s_{t})}
\label{importance_sampling_ratio}
\end{align}
is used to correct the distributional mismatch between the policies. This ratio adjusts the contribution of each sample in the update process to reflect how likely the action taken under the behavior policy $\pi_{b}$ would have been under $\pi$.

\subsection{Adversarial Inverse Reinforcement Learning}
IRL seeks to infer the reward function from the demonstration data $\mathcal{D}^{\star}$, which is interpreted as addressing the maximum likelihood problem 
\begin{align*}
\max_{\phi}\mathbb{E}_{\tau \sim \mathcal{D}^{\star}}\left[ \log p_{\phi}(\tau) \right], 
\end{align*}
where $p_{\phi}(\tau) \propto p(s_{0})\prod_{t} p\left(s_{t+1}|s_t, a_t\right) e^{\gamma^{t} r_{\phi}\left(s_{t}, a_{t}\right)}$. Building on the maximum causal entropy IRL framework \citep{ziebart2010modeling}, AIRL leverages a generative adversarial network (GAN) formulation \citep{goodfellow2014generative} to obtain solutions for this problem. To learn disentangled rewards that are invariant to changing dynamics, AIRL formalizes the discriminator as
\begin{align*}
D_{\phi, \Phi}\left(s, a, s'\right)=\frac{\exp \left\{f_{\phi,\Phi}\left(s, a, s'\right)\right\}}{\exp \left\{f_{\phi,\Phi}\left(s, a, s'\right)\right\}+\pi_{\theta}(a|s)},
\end{align*}
where $f_{\phi,\Phi}\left(s, a, s'\right)=g_{\phi}(s, a)+\gamma h_{\Phi}\left(s'\right)-h_{\Phi}(s)$, $g_{\phi}$ is a reward approximator, $h_{\Phi}$ is a shaping term and $\pi_{\theta}$ is the learned policy. Via binary logistic regression, AIRL trains $D_{\phi, \Phi}$ to classify expert data from policy samples. AIRL sets the reward as 
\begin{align}
r_{\phi, \Phi}\left(s, a, s'\right)=\log D_{\phi, \Phi}\left(s, a, s'\right)-\log(1-D_{\phi, \Phi}\left(s, a, s'\right))
\label{AIRL_reward}
\end{align}
and updates the policy $\pi_{\theta}$ by any policy optimization method. If the ground truth reward $r_{gt}$ only depends on the state, AIRL restricts the reward class to state-only rewards \citep{amin2017repeated} and is identified up to a constant, i.e., $g_{\phi}^{\star}(s)=r_{gt}(s)+const$. 

\subsection{Notation on Random Matrix Theory}
Henceforth, we denote $\lambda_{|\mathcal{S}|}(A)\leq \cdots \leq \lambda_1(A)$ as the ordered eigenvalues of one $|\mathcal{S}|\times |\mathcal{S}|$ Hermitian matrix $A$. The empirical spectral distribution (ESD) of $A$ is
\begin{eqnarray}\label{ESD}
F_{|\mathcal{S}|}(x):=\frac{1}{|\mathcal{S}|}\sum_{j=1}^{|\mathcal{S}|} {\mathbb I}_{\{\lambda_{j}(A) \leq x\}},\quad x\in\mathbb{R}.
\end{eqnarray}
Here and throughout the paper, ${\mathbb I}_{A}$ represents the indicator function of event $A$. The Stieltjes transform of $F_{|\mathcal{S}|}$ is given by
\begin{eqnarray*}
m_{|\mathcal{S}|}(z):=\int \frac{1}{x-z}{\rm d}F_{|\mathcal{S}|}(x),
\end{eqnarray*}
where $z=E+i\eta \in \mathbb{C}^{+}$.

For two positive quantities $A_n$ and $B_n$ that depend on $n$, we use the notation $A_n\asymp B_n$ to mean that $C^{-1}A_n\leq B_n\leq CA_n$ for some positive constant $C>1$. We need the following probability comparison definition from \citep{erdHos2013averaging}.

Let $\mathcal{X}\equiv \mathcal{X}^{(|\mathcal{S}|)}$ and $\mathcal{Y}\equiv \mathcal{Y}^{(|\mathcal{S}|)}$ be two sequences of nonnegative random variables. We say that $\mathcal{Y}$ stochastically dominates $\mathcal{X}$ if, for all (small) $\epsilon>0$ and (large) $D>0$,
\begin{eqnarray}
\mathbb{P}(\mathcal{X}^{(|\mathcal{S}|)}>|\mathcal{S}|^\epsilon\mathcal{Y}^{(|\mathcal{S}|)})\leq |\mathcal{S}|^{-D},
\label{intro9}
\end{eqnarray}
for sufficiently large $|\mathcal{S}|>|\mathcal{S}|_0(\epsilon,D)$, and we write $\mathcal{X}\prec\mathcal{Y}$ or $\mathcal{X}=O_\prec(\mathcal{Y})$. Moreover, if for some complex family $\mathcal{X}$ we use $\mathcal{X}\prec\mathcal{Y}$ to indicate $|\mathcal{X}|\prec\mathcal{Y}$, and we similarly write $\mathcal{X}=O_\prec(\mathcal{Y})$. When $\mathcal{X}^{(|\mathcal{S}|)}$ and $\mathcal{Y}^{(|\mathcal{S}|)}$ depend on a parameter $v\in \mathcal{V}$ (typically an index label or a spectral parameter), then $\mathcal{X}\prec\mathcal{Y}$ uniformly on $v\in \mathcal{V}$, which means that the threshold $|\mathcal{S}|_0(\epsilon,D)$ can be chosen independently of $v$. Next, we say that an event $\Xi=\Xi^{|\mathcal{S}|}$ holds with high probability if $1-{\mathbb I}(\Xi) \prec 0$, i.e., if for any $D>0$ there is $|\mathcal{S}|_0(D)$ such that for all $|\mathcal{S}|\geq |\mathcal{S}|_0(D)$ we have $\mathbb P (\Xi^{|\mathcal{S}|})\geq 1-|\mathcal{S}|^{-D}$. We use the symbols $O(\cdot)$ and $o(\cdot)$ for the standard big-O and little-o notations, respectively. We use $c$ and $C$ to denote strictly positive constants that do not depend on $|\mathcal{S}|$. Their values may change from line to line. For any matrix $A$, we denote $\|A\|$ as its operator norm, while for any vector $\bm v$, we use $\|\bm v\|$ to denote its $L_2$-norm. 

\section{Transferability Condition on Environment Dynamics\label{sec_disentangled_condition_environment_dynamics}}
We first recall disentangled rewards, which are invariant to changing dynamics. 
\begin{definition}[Disentangled rewards \citep{fu2018learning}]
A reward function $r'(s,a,s')$ is (perfectly) disentangled with respect to a ground truth reward $r_{gt}(s,a,s')$ and a set of dynamics $\mathcal{P}$ such that under all dynamics $p\in \mathcal{P}$, the optimal policy is the same: $\pi_{r',p}^{\star}(a|s)=\pi_{r_{gt},p}^{\star}(a|s)$. 
\end{definition}

Recall the relationship between the optimal policy and its corresponding Q-function\footnote{In this version, we adopt the assumption of a single optimal action associated with each state to simplify the analysis. Importantly, this assumption does not compromise the validity of our findings; even if multiple actions yield the same optimal Q-function value in practice, it does not influence our analytical results.}\citep{sutton2018reinforcement}: 
\begin{align*}
\pi_{r',p}^{\star}(a|s)
=
\begin{cases}
1&,~({\rm if}~a=\mathop{\arg\max}_{a\in \mathcal{A}}Q_{r',p}^{\star}(s,a)) \\
0&,~~~~~~~~~~~~~~~~~~~({\rm else})
\end{cases}
, \\
\pi_{r_{gt},p}^{\star}(a|s)
=
\begin{cases}
1&,~({\rm if}~a=\mathop{\arg\max}_{a\in \mathcal{A}}Q_{r_{gt},p}^{\star}(s,a)) \\
0&,~~~~~~~~~~~~~~~~~~~({\rm else})
\end{cases}
. 
\end{align*}
Therefore, $\pi_{r',p}^{\star}(a|s)$ and $\pi_{r_{gt},p}^{\star}(a|s)$ being equal is equivalent to their corresponding Q-functions up to arbitrary action-independent functions $f(s)$, i.e., $Q_{r',p}^{\star}(s,a)=Q_{r_{gt},p}^{\star}(s,a)-f(s)$. 

In the following, we analyze the transferability condition on environment dynamics from a rank perspective. 
\begin{theorem}\label{theorem_disentangled_rank_condition}
Let $r_{gt}(s)$ be a ground truth reward, $p$ is a dynamics model, and $\pi_{p}^{\star}$ is the optimal policy under $r_{gt}$ and $p$. Suppose $r'(s)$ is the reward recovered by AIRL that produces an optimal policy $\pi_{p}^{\star}$ in $p$:
\begin{align*}
Q_{r',p}^{\star}(s,a)=Q_{r_{gt},p}^{\star}(s,a)-f(s). 
\end{align*}
If ${\rm rank}(\gamma \mathbf{P}-I)=|\mathcal{S}|-1$, when $\gamma$ approaches $1$, i.e., ${\rm rank}(\mathbf{P}-I)=|\mathcal{S}|-1$, then $r'(s)$ is disentangled with respect to all dynamics. 
\end{theorem}

The choice of setting $\gamma$ to approach 1 is commonly adopted in RL \citep{puterman2014markov, sutton2018reinforcement, kurutach2018model, gottesman2023coarse}, as it effectively places a greater emphasis on long-term rewards over immediate ones, aligning the agent's decision-making process with the objective of maximizing cumulative future rewards. This approach ensures that the agent remains focused on learning strategies that perform well over extended time horizons, which is often desirable in many RL applications. We refer to 
\begin{align}
{\rm rank}(\mathbf{P}-I)=|\mathcal{S}|-1,
\label{transferabilitycondition}
\end{align}
as the \textit{transferability condition}.

Before proving Theorem \ref{theorem_disentangled_rank_condition}, we first introduce an important lemma. 

\begin{lemma}\citep{fu2018learning}
Let $r_{gt}(s)$ be a ground truth reward, $p$ is a dynamics model, and $\pi_{p}^{\star}$ is the optimal policy under $r_{gt}$ and $p$. Suppose $r'(s)$ is the reward recovered by AIRL that produces an optimal policy $\pi_{p}^{\star}$ in $p$:
\begin{align*}
Q_{r',p}^{\star}(s,a)=Q_{r_{gt},p}^{\star}(s,a)-f(s). 
\end{align*}
Then
\begin{align}
r'(s)=r_{gt}(s)+\gamma \mathbb{E}_{s'\sim p}\left[ f(s') \right]-f(s). 
\label{r'_rgt_expectation_relationship}
\end{align}
\label{lemma_reward_shaping}
\end{lemma}

\textbf{Proof of Theorem \ref{theorem_disentangled_rank_condition}:}
\begin{proof}
Assume that for some arbitrary state-dependent function $b(s)$, $r'(s)=r_{gt}(s)+b(s)$. By Lemma \ref{lemma_reward_shaping}, we have that for all $s,a$, $b(s)=\gamma \mathbb{E}_{s'\sim p}\left[ f(s') \right]-f(s)$, 
i.e., 
\begin{align}
b(s)=\gamma \sum_{s'}p(s'|s,a)f(s')-f(s). 
\label{b_wrt_s_a}
\end{align}
Taking expectations for the optimal policy $\pi_{p}^{\star}$ on both sides of \eqref{b_wrt_s_a}, we derive that
\begin{align}
b(s)=\gamma \sum_{s'}p(s'|s)f(s')-f(s),
\label{system}
\end{align}

\eqref{system} forms a non-homogeneous system of linear equations in terms of $f$. Let $I$ be the identity matrix, $X=[f(s^{1}),\ldots,f(s^{|\mathcal{S}|})]^{\top}$, $b=[b(s^{1}),\ldots,b(s^{|\mathcal{S}|})]^{\top}$. Then we obtain that 
\begin{align*}
\left( \gamma \mathbf{P}-I \right) X=b. 
\end{align*}

If ${\rm rank}(\gamma \mathbf{P}-I)=|\mathcal{S}|-1$, then the homogeneous system of linear equations
\begin{align}
\left( \gamma \mathbf{P}-I \right) X=0 
\label{matrix_equation_no_b}
\end{align}
has one free variable. 

When $\gamma$ approaches 1, i.e., transferability condition \eqref{transferabilitycondition} holds, the fact $\sum_{j} p(s^{j}|s^{i}) = 1$ guarantees that \eqref{matrix_equation_no_b} has the solutions approaching $c[1,\ldots,1]^{\top}$ for any constant $c$. At this moment, $f(s)$ is a constant, so by \eqref{system}, we derive that $b(s)$ is a constant. Consequently, $r'(s)$ equals the ground truth reward $r_{gt}(s)$ up to a constant, thereby inducing the same optimal policy and completing the proof of Theorem \ref{theorem_disentangled_rank_condition}. Note that in our proof, the transferability condition \eqref{transferabilitycondition} and the assertion that $f(s)$ is a constant are both necessary and sufficient.
\end{proof}

\begin{remark}
In \citep[Theorem 5.1]{fu2018learning}, it is directly derived that $f(s)$ is constant by applying \eqref{r'_rgt_expectation_relationship} under the decomposability condition. 

``Under the decomposability condition, every state in the MDP can be linked with such an equality statement, meaning that $f(s)$ is constant.'' 

However, the decomposability condition — which states that functions over current states $f(s)$ and next states $g(s')$ can be separated from their sum $f(s) + g(s')$ — is overly restrictive and not intuitive, making it difficult to verify.

An alternative detailed explanation is provided in Theorem \ref{theorem_disentangled_rank_condition}, which is based on the rank condition \eqref{transferabilitycondition}.
\end{remark}

In practice, the transition matrix $\mathbf{P}$ of the state space is an unknown matrix, but we can prove that the transferability condition ${\rm rank}(\mathbf{P}-I)=|\mathcal{S}|-1$ holds naturally and the transferability remains robust as $|\mathcal{S}| \to \infty$ (characterized by exactly one singular value of $\mathbf{P}-I$ equaling 0, while the remaining $|\mathcal{S}|-1$ singular values are significantly non-zero). From the perspective of variational inference (Bayesian statistics) \citep{efron2016computer,levine2018reinforcement,hu2023sampling}, we first establish this rank condition under an uninformative prior when no prior information is available. We then extend the proof to scenarios with informative priors, such as known obstacle locations in the environment, where specific elements of $\mathbf{P}$ are $0$.

\subsection{Uninformative Prior on $\mathfrak{P}$}
Recall the probability model $\mathfrak{P}=\left(p_{i j}\right) \in \mathbb{R}^{|\mathcal{S}| \times|\mathcal{S}|}$ of transition matrix $\mathbf{P}$, derived using variational inference (Bayesian statistics) \citep{efron2016computer,levine2018reinforcement} when the true matrix $\mathbf{P}$ is unobservable. In the absence of prior information, we assume that $\mathfrak{P}$ follows an uninformative prior distribution, known as flat Dirichlet distribution. Specifically, $p_{ij}=x_{ij} /\left(\sum_{j=1}^{|\mathcal{S}|} x_{i j}\right)$, where $x_{ij}$'s are i.i.d. random variables following the one-sided exponential distribution with expectation $1$ and density function $f(x)=e^{-x}$ for $x \in[0, \infty)$.

Denote $W:= \mathfrak{P} - I$. In the large dimension framework
$$
|\mathcal{S}| \rightarrow \infty,
$$
to prove that $\text{rank}(\mathfrak{P}-I) = |\mathcal{S}| - 1$ with high probability, it suffices to demonstrate two points: 
\begin{itemize}
\item[(i)] there are linear relationships among the columns of $W$;
\item[(ii)] with high probability, at least $|\mathcal{S}| - 1$ singular values of $W$ are far from 0.
\end{itemize}

The first point is guaranteed by the equation
\begin{align}\label{hanghe0}
\sum_{j=1}^{|\mathcal{S}|} p_{ij} - 1 = 0,
\end{align}
for all $i = 1, \ldots, |\mathcal{S}|$, which ensures $\text{rank}(W) \leq |\mathcal{S}| - 1$. The second point establishes that $\text{rank}(W) \geq |\mathcal{S}| - 1$. Hereafter we use $s_{|\mathcal{S}|}(A) \leq \ldots \leq s_{1}(A)$ to denote the ordered singular values of matrix $A$. We will now provide detailed proof for point (ii).

\begin{theorem}\label{mainth}
For the probability model $W = \mathfrak{P} - I$, we have the following two statements:
\begin{enumerate}
\item $s_{|\mathcal{S}|}(W) = 0$;
\item For all $j = 1, \ldots, |\mathcal{S}| - 1$, the singular values $s_j(W)$ satisfy $|s_j(W) - 1| \prec |\mathcal{S}|^{-1/4}$.
\end{enumerate}
\end{theorem}

Statement 1 $s_{|\mathcal{S}|}(W) = 0$ follows from \eqref{hanghe0}. We only need to prove statement 2. The following lemma from \citep{beh2016} collects properties of stochastic domination $\prec$. Roughly, it states that $\prec$ satisfies the usual arithmetic properties of order relations. We shall use it tacitly throughout the following.

\begin{lemma}\label{Basic properties of}(Lemma 3.4 in \citep{beh2016})
\begin{itemize}
\item[(i)] Suppose that $X(u, v) \prec Y(u, v)$ uniformly on $u \in U$ and $v \in V$. If $|V| \leq |\mathcal{S}|^{C}$ for some constant $C$, then
\begin{align*}
\sum_{v \in V} X(u, v) \prec \sum_{v \in V} Y(u, v)
\end{align*}
uniformly on $u$.

\item[(ii)] Suppose that $X_1(u) \prec Y_1(u)$ uniformly on $u$ and $X_2(u) \prec Y_2(u)$ uniformly on $u$. Then
\begin{align*}
X_1(u) X_2(u) \prec Y_1(u) Y_2(u)
\end{align*}
uniformly on $u$.
\end{itemize}
\end{lemma}

\textbf{Proof of Theorem \ref{mainth}:}
\begin{proof}
Define $b_{ij}=p_{ij}-|\mathcal{S}|^{-1}$ for all $1\leq i,j \leq |\mathcal{S}|$. We can then express $b_{ij}$ as follows:
\begin{align}
b_{ij}&=\frac{x_{ij}}{\sum_{j=1}^{|\mathcal{S}|} x_{i j}}-|\mathcal{S}|^{-1} \notag\\
&=|\mathcal{S}|^{-1}\left(\frac{x_{ij}}{|\mathcal{S}|^{-1}\sum_{j=1}^{|\mathcal{S}|}x_{ij}}-1\right) \notag\\
&=|\mathcal{S}|^{-1}\bar{x}_{i}^{-1}(x_{ij}-\bar{x}_i),
\end{align}
where $\bar{x}_i=|\mathcal{S}|^{-1}\sum_{j=1}^{|\mathcal{S}|}x_{ij}$ represents the sample mean for $x_{ij},~j=1,\ldots, |\mathcal{S}|$.
For the asymmetric $B$, we symmetrize it as
\begin{align*}
\tilde{b}_{ij}
=
\begin{cases}
b_{ij}&,~~~~~~({\rm for}~~i\leq j) \\
b_{ji}&.~~~~~~({\rm for}~~i> j)
\end{cases}
\end{align*}

Now, define
\begin{align}
q_{ij}:=|\mathcal{S}|^{-1/2}(x_{ij}-\bar{x}_i),
\label{qij}
\end{align}
\begin{align}
\tilde{q}_{ij}:=|\mathcal{S}|^{-1/2}\bar{x}_{i}^{-1}(x_{ij}-\bar{x}_i),
\label{tidleqij}
\end{align}
where the matrices corresponding to $q_{ij}$ and $\tilde{q}_{ij}$ are denoted by $Q:=(q_{ij})\in \mathbb{R}^{|\mathcal{S}| \times|\mathcal{S}|}$ and $ \tilde{Q}:=(\tilde{q}_{ij})\in \mathbb{R}^{|\mathcal{S}| \times|\mathcal{S}|}$, respectively. Similarly, the matrix corresponding to $b_{ij}$ and $\tilde{b}_{ij}$ are denoted by $B:=(b_{ij})\in \mathbb{R}^{|\mathcal{S}|\times|\mathcal{S}|}$ and $\tilde{B}:=(\tilde{b}_{ij})\in \mathbb{R}^{|\mathcal{S}|\times|\mathcal{S}|}$, respectively. We then have
\begin{align}
\tilde{Q}:=D^{-1}Q,
\label{Qmatrix}
\end{align}
where $D$ is a diagonal matrix with $\bar{x}_{i}$ as its $i$-th diagonal element, i.e., 
\begin{align}
D=\text{Diag}\{\bar{x}_{1},\ldots,\bar{x}_{|\mathcal{S}|}\}.
\label{D}
\end{align}
Furthermore, we have the relation
\begin{align}
B=|\mathcal{S}|^{-1/2}\tilde{Q}. 
\label{Q&B}
\end{align}
Next, consider the difference between $Q$ and $\tilde{Q}$. We obtain the bound by 
\begin{align}
|s_i(Q)-s_i(\tilde{Q})|\leq \|\tilde{Q}- Q\|\leq \|D^{-1}-I\| \cdot \|Q\|\leq \|D^{-1}\| \cdot \|D-I\| \cdot \|Q\|.
\label{Q difference}
\end{align}
Now we claim that 
\begin{align}
\|D-I\|\prec |\mathcal{S}|^{-1/2}.
\label{D-IBOUND}
\end{align}
This bound follows from the estimate
\begin{align}
\mathbb P(\|D-I\|\geq |\mathcal{S}|^{-1/2+\epsilon})
&\leq |\mathcal{S}|\mathbb P(|\bar{x}_1-1|\geq |\mathcal{S}|^{-1/2+\epsilon}) \notag\\
&\leq |\mathcal{S}|^{-2q\epsilon+q+1} \mathbb E (\bar{x}_1-1)^{2q} \notag\\
&\leq C_q|\mathcal{S}|^{-2q\epsilon+1},
\label{profD-I}
\end{align}
for $q>1$. From this, we can also derive
\begin{align*}
\mathbb P(\max_{i=1}^{|\mathcal{S}|}\{|\bar{x}_{i}-1|\}\geq |\mathcal{S}|^{-1/2+\epsilon})\leq C_q|\mathcal{S}|^{-2q\epsilon+1},
\end{align*}
which implies
\begin{align}\label{maxxi}
\max_{i=1}^{|\mathcal{S}|}\{|\bar{x}_{i}-1|\}\prec|\mathcal{S}|^{-1/2}.
\end{align}
Combining this with $\lambda_{1}(D^{-1})=\lambda^{-1}_{|\mathcal{S}|}(D)$, we conclude that
\begin{align}
\|D^{-1}\|\prec C
\label{D-bound}
\end{align}
for some large constant $C$.

Define the empirical spectral distribution (ESD) of $\tilde{B}$ and $\tilde{B} + |\mathcal{S}|^{-1} ee^{\top}$ as follows:
\begin{align*}
F^{\tilde{B}}(x)=\frac{1}{|\mathcal{S}|}\sum_{j=1}^{|\mathcal{S}|}{\mathbb I}_{\{\lambda_{j}(\tilde{B}) \leq x\}},\quad x\in\mathbb{R},\\
F^{\tilde{B} + |\mathcal{S}|^{-1} ee^{\top}}(x)=\frac{1}{|\mathcal{S}|}\sum_{j=1}^{|\mathcal{S}|}{\mathbb I}_{\{\lambda_{j}(\tilde{B}+|\mathcal{S}|^{-1} ee^{\top}) \leq x\}},\quad x\in\mathbb{R},
\end{align*}
where $e = (1, \ldots, 1)^{\top}$ is a $|\mathcal{S}|$-dimensional vector with all elements equal to 1.

Based on the preceding analysis, we can prove Statement 2 of Theorem \ref{mainth} using the following three lemmas.

\begin{lemma}\label{Perturbation}
Under the same setting as Theorem \ref{mainth}, we have: $$F^{\tilde{B}}(x)-F^{\tilde{B} + |\mathcal{S}|^{-1} ee^{\top}}(x)\leq |\mathcal{S}|^{-1}.$$
\end{lemma} 

\begin{lemma}\label{maxeig}
Under the same setting as Theorem \ref{mainth}, we have:
$$\|B-\tilde{B}\|\prec |\mathcal{S}|^{-1/4}.$$
\end{lemma}

\begin{lemma}\label{bound}
Under the same setting as Theorem \ref{mainth}, we have: $$|\lambda_1(QQ^{\top})-4|\prec |\mathcal{S}|^{-2/3},$$ and then $$\lambda_k(QQ^{\top})\prec C,$$ for some large constant $C$ for all $k=1,\ldots,|\mathcal{S}|$.
\end{lemma}

The proof of Lemma \ref{Perturbation}, Lemma \ref{maxeig} and Lemma \ref{bound} are deferred to Supplementary Material. We now proceed to prove Theorem \ref{mainth}.

Using Lemma \ref{bound} along with \eqref{Q difference}, \eqref{D-IBOUND} and \eqref{D-bound}, we derive that $s_i(\tilde{Q}) \prec C$. Combining this with \eqref{Q&B} leads to $s_k(B) \prec |\mathcal{S}|^{-1/2}$ for $k = 1, \ldots, |\mathcal{S}|$. By applying Lemma \ref{maxeig}, we conclude that 
\begin{align}\label{Bmin}
s_k(\tilde{B}) \prec |\mathcal{S}|^{-1/4} 
\end{align} 
for all $k = 1, \ldots, |\mathcal{S}|$.

Next, by applying Lemma \ref{Perturbation} and \eqref{Bmin}, we have
\begin{align*}
s_k(\tilde{B}+|\mathcal{S}|^{-1} ee^{\top})\prec |\mathcal{S}|^{-1/4}
\end{align*}
uniformly for at least $|\mathcal{S}| - 1$ singular values (and similarly for eigenvalues, due to the symmetry of $\tilde{B} + |\mathcal{S}|^{-1} ee^{\top}$). Thus, the corresponding singular values of $\tilde{B}+ |\mathcal{S}|^{-1} ee^{\top}-I$ satisfy
\begin{align*}
|s_k(\tilde{B}+|\mathcal{S}|^{-1} ee^{\top}-I)-1|\prec |\mathcal{S}|^{-1/4}.
\end{align*}
Furthermore, applying Lemma \ref{maxeig} again yields
\begin{align}\label{w-b} 
\nonumber |s_k(W) - s_k(\tilde{B} + |\mathcal{S}|^{-1} ee^{\top} - I)| &= |s_k(\mathfrak{P} - I) - s_k(\tilde{B} + |\mathcal{S}|^{-1} ee^{\top} - I)| \\
\nonumber &= |s_k(B + |\mathcal{S}|^{-1} ee^{\top} - I) - s_k(\tilde{B} + |\mathcal{S}|^{-1} ee^{\top} - I)| \\
&\leq \|B - \tilde{B}\| \prec |\mathcal{S}|^{-1/4}. 
\end{align}
Consequently, from \eqref{w-b}, we conclude that ${\rm rank}(W) \geq |\mathcal{S}| - 1$ with high probability. This, combined with Lemma \ref{maxeig}, proves Theorem \ref{mainth}.
\end{proof}

\begin{remark}
We employ the Dirichlet distribution (a common setting in Bayesian statistics) solely as an illustrative example. It is not a requirement for the $x_{ij}$ to adhere to an exponential distribution; indeed, each $x_{ij}$ can be governed by a distinct distribution. The only condition is that they must satisfy:
\begin{align*}
\lim_{s \to \infty} s^4 \mathbb{P}\left(|\mathcal{S}|^{-1/2}|x_{ij}| \leq s\right) = 0,
\end{align*}
for $1\leq i,j \leq |\mathcal{S}|$.
For further insights from RMT, refer to \citep{ding2018necessary}.
\end{remark}

\subsection{Informative Prior on $\mathfrak{P}$}
When certain elements of $\mathfrak{P}$ are known to be $0$ (representing impassable obstacle locations), we assume, without loss of generality, that obstacles are present near the $k$-th state $s^{k}$. This implies that the $k$-th row of $\mathfrak{P}$ contains $(1-\omega)|\mathcal{S}|$ zeros, while the remaining $\omega|\mathcal{S}|$ elements follow an uninformative prior. Specifically, for $j = 1, \ldots, \omega|\mathcal{S}|$, we have $p_{k j} = x_{k j}/\left(\sum_{j=1}^{\omega|\mathcal{S}|} x_{k j}\right)$, where $x_{k j}$ are i.i.d. random variables drawn from a one-sided exponential distribution with mean $1$ and density function $f(x) = e^{-x}$ for $x \in [0, \infty)$. For $j = \omega|\mathcal{S}| + 1, \ldots, |\mathcal{S}|$, we set $p_{k j} = x_{k j} = 0$. Let $\mathfrak{P}_1$ denote this new probability model with the informative prior. For convenience, we assume $\omega|\mathcal{S}|$ is an integer.

Recalling the definitions of $B$, $Q$ and $\tilde{Q}$, the obstacle locations near the $k$-th state-only alter the $k$-th row of $B$, which is updated as follows:
\begin{align*}
b^1_{kj}&=\frac{x_{kj}}{\sum_{j=1}^{\omega|\mathcal{S}|} x_{k j}}-(\omega|\mathcal{S}|)^{-1} \notag\\
&=(\omega|\mathcal{S}|)^{-1}\left(\frac{x_{kj}}{(\omega|\mathcal{S}|)^{-1}\sum_{j=1}^{|\omega\mathcal{S}|}x_{kj}}-1\right) \notag\\
&=(\omega|\mathcal{S}|)^{-1}\bar{x}_{k}^{-1}(x_{kj}-\bar{x}_k),~~~j=1,\ldots, \omega|\mathcal{S}|,\\
b^1_{kj}&=0,~~~j=\omega|\mathcal{S}|+1,\ldots,|\mathcal{S}|,
\end{align*}
where $\bar{x}_k=(\omega|\mathcal{S}|)^{-1}\sum_{j=1}^{\omega|\mathcal{S}|}x_{kj}$ represents the sample mean for $x_{ij},~j=1,\ldots, \omega|\mathcal{S}|$. We also update $Q$ as follows:
\begin{align*}
q^1_{kj}&=|\mathcal{S}|^{-1/2}(x_{kj}-\bar{x}_k),~~~j=1,\ldots, \omega|\mathcal{S}|,\\
q^1_{kj}&=0,~~~j=\omega|\mathcal{S}|+1,\ldots,|\mathcal{S}|.
\end{align*}
Similarly, $\tilde{Q}$ is updated as:
\begin{align*}
\tilde{q}^1_{kj}&=\omega^{-1}|\mathcal{S}|^{-1/2}\bar{x}_{k}^{-1}(x_{kj}-\bar{x}_k),,~~~j=1,\ldots, \omega|\mathcal{S}|,\\
\tilde{q}^1_{kj}&=0,~~~j=\omega|\mathcal{S}|+1,\ldots,|\mathcal{S}|.
\end{align*}
The updated matrices are denoted as $B_1$, $Q_1$ and $\tilde{Q}_1$. Revisiting equation \eqref{Q difference}, we have:
$$
|s_i(Q_1) - s_i(\tilde{Q}_1)| \leq \|\tilde{Q}_1 - Q_1\| \leq \|D_1^{-1} - I\| \cdot \|Q_1\| \leq \|D_1^{-1}\| \cdot \|D_1 - I\| \cdot \|Q_1\|,
$$
where $D_1$ is obtained by replacing the $k$-th element of $D$ with $\omega^{-1}\bar{x}_k$.

It directly follows that:
$$
\|D_1 - I\| \prec \max\{C,\omega^{-1}\},
$$
and
$$
\|D_1^{-1}\| \prec C.
$$

By applying Lemma \ref{lemma2}, Theorem \ref{mainth} can be extended to $\mathfrak{P}_1$ with $\omega \in (0, 1]$, utilizing the same proof strategy as in the uninformative prior case.

\begin{lemma}\label{lemma2}
For $\omega \in (0, 1]$, we have $\lambda_k(Q_1 Q_1^{\top}) \prec C$ for some large constant $C$ for all $k = 1, \ldots, |\mathcal{S}|$. \end{lemma}

Lemma \ref{lemma2} directly follows from 
\begin{align}
|s_k(Q) - s_k(Q_1)| \leq \|Q-Q_1\|=\sqrt{\lambda_1\left((Q-Q_1)(Q-Q_1)^{\top}\right)}\prec C,
\label{sq-sq1}
\end{align}
for some large constant $C$.

\begin{remark} 
The above analysis and Lemma \ref{lemma2}, established via \eqref{sq-sq1}, can be extended to the finite-rank case; more generally, the rank can approach $|\mathcal{S}|^{\epsilon}$ for small $\epsilon$. This extension allows the environment to incorporate obstacles of finite rank or other informative priors.
\end{remark}

The analyses in this section reframe AIRL by our rank condition, which suggests that the problem of inadequate transfer stems from an external factor rather than the framework itself (the performance issues are not due to its design). In the following section, we will elaborate on this point to strengthen the practical persuasiveness of AIRL.

\section{Reward Transferability Analysis\label{sec_reward_transferable_analysis}}
Based on our analysis that AIRL's transferability condition on environment dynamics is naturally satisfied during training, we shift our focus to another factor affecting transfer effectiveness: the selection of the RL algorithm employed by AIRL, specifically the choice between on-policy and off-policy methods. We first identify the ineffective reward extraction by off-policy RL algorithm in the source environment (Section \ref{sec_extractability_of_disentangled_rewards}). Next, we highlight the effectiveness of the off-policy RL algorithm during the policy re-optimization in the target environment (based on the extracted reward) and introduce our hybrid framework PPO-AIRL + SAC (Section \ref{sec_PPO-AIRL + SAC}).

\subsection{Employing an On-policy or Off-policy RL Algorithm in AIRL?\label{sec_extractability_of_disentangled_rewards}}
Our following finding shows that employing an off-policy RL algorithm in AIRL leads to ineffective reward extraction in the source environment.

Let $w$ be the parameter of the Q-function. Recall the importance sampling ratio $\rho_{t}$ (\eqref{importance_sampling_ratio}) within off-policy RL algorithm in Section \ref{sec_MDP}, the quantity being updated can be formulated by $\Delta w=\alpha \rho_{t}\delta_{t}\nabla_{w}Q_{w}(s_{t},a_{t})$ during the source training process, where $\alpha$ is the learning rate and $\delta_{t}=r_{t+1}+\gamma \max_{a}Q_{w}(s_{t+1},a)-Q_{w}(s_{t},a_{t})$ \citep{sutton2018reinforcement}. Then the variance of $\Delta w$ is given by:
\begin{align*}
\text{Var}_{\pi_b}[\Delta w]
&=\text{Var}_{\pi_b} \left( \alpha \rho_{t}\delta_{t}\nabla_{w}Q_{w}(s_{t},a_{t}) \right) \\
&=\alpha^{2}\text{Var}_{\pi_b} \left( \rho_{t}\delta_{t}\nabla_{w}Q_{w}(s_{t},a_{t}) \right). 
\end{align*}
Next, applying the variance of products formula: 
\begin{align*}
\text{Var}[XY] = \mathbb{E}[X]^2 \text{Var}[Y] + \mathbb{E}[Y]^2 \text{Var}[X] + \text{Var}[X] \text{Var}[Y]
\end{align*}
with $X=\rho_{t}$, we get that
\begin{align*}
\text{Var}_{\pi_b}[\Delta w] = \alpha^{2} \left( \mathbb{E}_{\pi_b}[\rho_{t}]^{2} \text{Var}_{\pi_b}[Y] + \mathbb{E}_{\pi_b}[Y]^{2} \text{Var}_{\pi_b}[\rho_{t}] + \text{Var}_{\pi_b}[\rho_{t}] \text{Var}_{\pi_b}[Y] \right), 
\end{align*}
where $Y=\delta_{t}\nabla_{w}Q_{w}(s_{t},a_{t})$. 

Note that the variance of $\rho_{t}$ depends on the difference between policy $\pi$ and behavior policy $\pi_b$. Since $\rho_{t} = \pi(a_t|s_t)/\pi_b(a_t|s_t)$, its variance grows as the two policies are not aligned. The expectation and variance of $\rho_{t}$ are: 
\begin{align*}
\mathbb{E}_{\pi_b}[\rho_{t}] = \mathbb{E}_{\pi_b}\left[\frac{\pi(a_t|s_t)}{\pi_b(a_t|s_t)}\right] = \sum_{a_t} \pi(a_t|s_t)=1. 
\end{align*}
\begin{align*}
\text{Var}_{\pi_b}[\rho_{t}] = \mathbb{E}_{\pi_b}[\rho_{t}^2] - \mathbb{E}_{\pi_b}[\rho_{t}]^2 = \mathbb{E}_{\pi_b}\left[\left( \frac{\pi(a_t|s_t)}{\pi_b(a_t|s_t)} \right)^2 \right] - 1. 
\end{align*}

For \textbf{off-policy RL algorithm}, a distribution shift exists between $\pi_b$ and $\pi$, which results in a positive variance, $\text{Var}_{\pi_b}[\rho_{t}]$. In contrast, for \textbf{on-policy RL algorithm}, $\text{Var}_{\pi_b}[\rho_{t}]=0$. This variance causes $\text{Var}_{\pi_b}[\Delta w]$ to be larger in off-policy optimization than in on-policy optimization, and it propagates through the update mechanism, leading to instability. 

Note that the reward recovery process in AIRL is ``meticulous''; it adjusts $\pi$ to mitigate variance during training by minimizing the Kullback–Leibler (KL) divergence:
\begin{align*}
D_{\rm KL}(\pi(\tau), p_{\phi}(\tau)), 
\end{align*}
where $\tau \sim \mathcal{D}^{\star}, \pi(\tau)=p(s_{0})\prod_{t} p\left(s_{t+1}|s_{t}, a_{t} \right) \pi \left( a_{t}|s_{t} \right)$ and 
\begin{align*}
p_{\phi}(\tau) \propto p(s_{0})\prod_{t} p\left(s_{t+1}|s_t, a_t\right) e^{\gamma^{t} r_{\phi}\left(s_{t}, a_{t}\right)}. 
\end{align*}

The large variance $\text{Var}_{\pi_b}[\Delta w]$ in off-policy RL algorithms contradicts the reward recovery principle of AIRL; in contrast, on-policy approaches are more suitable in the source training process. 

In the next subsection, we shift our focus to policy re-optimization in the target environment and present the global training structure: the hybrid framework PPO-AIRL + SAC.

\subsection{Hybrid Framework PPO-AIRL + SAC\label{sec_PPO-AIRL + SAC}}
The hybrid framework PPO-AIRL + SAC utilizes PPO-AIRL to recover the reward in the source environment and then applies SAC to re-optimize the policy (using the recovered reward) in the target environment. The strengths of PPO-AIRL + SAC arise from two key aspects: 
\begin{enumerate}
\item the precise and stable extraction of disentangled rewards through on-policy PPO-AIRL in the source environment, where meticulous reward recovery is essential;
\item the high sample efficiency of off-policy SAC in the target environment, enabling efficient and low-cost policy training.
\end{enumerate}
The entire procedure of PPO-AIRL + SAC is detailed in Algorithm \ref{alg_ppo_airl_sac}. 

\begin{algorithm}[htbp]
\caption{PPO-AIRL + SAC}
\begin{algorithmic}[1]
\STATE \textbf{Input:} Source expert demonstrations $\mathcal{D}^{\star}$. 
\STATE Initialize policy $\pi_{\theta}$ and discriminator $D_{\phi, \Phi}$ in the source environment. 
\STATE Train $D_{\phi, \Phi}$ via PPO-AIRL from $\mathcal{D}^{\star}$ in the source environment. 
\STATE Set reward $r_{\phi, \Phi}$ by \eqref{AIRL_reward}. 
\STATE Train the policy with respect to $r_{\phi, \Phi}$ by SAC in the target environment. 
\end{algorithmic}
\label{alg_ppo_airl_sac}
\end{algorithm}

\section{Experiments\label{sec_experiments}}
We first simulate the eigenvalue locations (Section \ref{sec_simulation_eigenvalues_location}), followed by verifying the transfer effect of PPO-AIRL + SAC (Section \ref{sec_transfer_effect_PPO-AIRL + SAC}).

\subsection{Simulation on Location of Eigenvalues\label{sec_simulation_eigenvalues_location}}
In this subsection, we will run simulations to validate the following rules:
\begin{itemize}
\item[(i)] Local eigenvalues $\lambda_i(QQ^{\top})$ exhibit rigidity (cluster near their typical location $\gamma_i$).
\item[(ii)] Only one singular value of $W=\mathfrak{P}-I$ equal $0$, while the remaining eigenvalues are located near $1$ with high probability, irrespective of whether the priors are uninformative or informative.
\end{itemize}

All simulations are performed with $|\mathcal{S}|=900$ and $|\mathcal{S}|=2500$. 

\begin{figure}[htbp]
\centering
\begin{minipage}{0.49\linewidth}
\vspace{1pt}
\includegraphics[width=\textwidth]{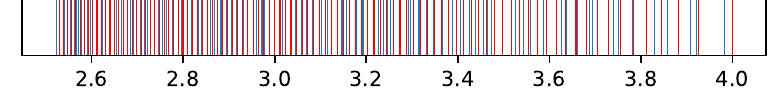}
\centerline{(a)}
\end{minipage}
\begin{minipage}{0.49\linewidth}
\vspace{1pt}
\includegraphics[width=\textwidth]{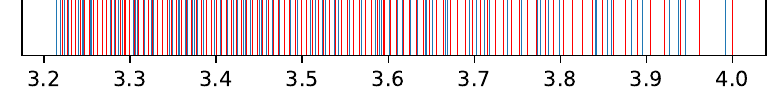}
\centerline{(b)}
\end{minipage}
\caption{Local eigenvalues locations for $QQ^{\top}$. (a) $|\mathcal{S}|=900$, (b) $|\mathcal{S}|=2500$. }
\label{fig_local_eigenvalues_locations_QQT}
\end{figure}

We first present the local eigenvalue locations of $QQ^{\top}$. We exhibit the top 100 eigenvalues in Fig.\,\ref{fig_local_eigenvalues_locations_QQT}. The blue and red lines represent the eigenvalues $\lambda_i(QQ^{\top})$ and typically locations $\gamma_{i}$, respectively. It can be observed that the eigenvalues $\lambda_i(QQ^{\top})$ are located near their corresponding $\gamma_{i}$ (with $\gamma_{1}=4$) for both $|\mathcal{S}|=900$ and $|\mathcal{S}|=2500$.

\begin{figure}[htbp]
\centering
\begin{minipage}{0.49\linewidth}
\vspace{1pt}
\includegraphics[width=\textwidth]{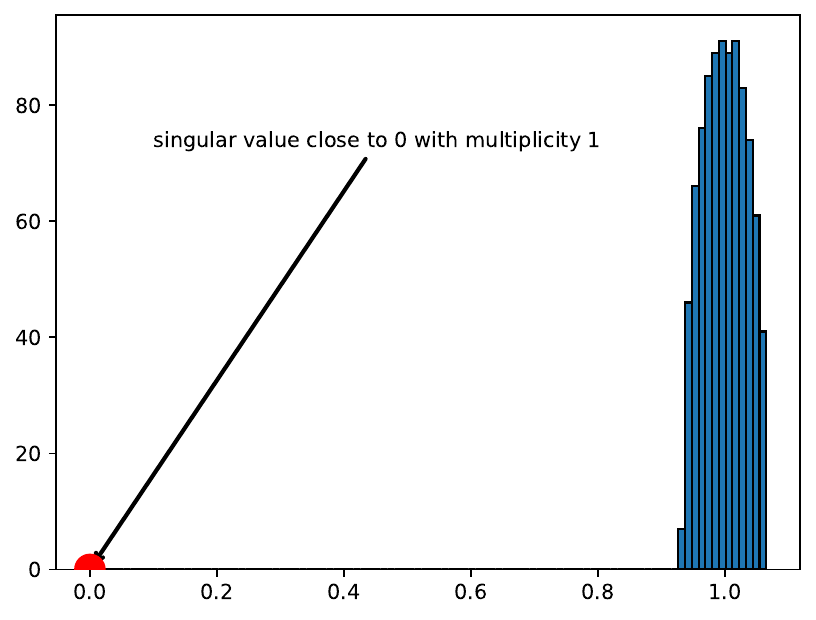}
\centerline{(a)}
\end{minipage}
\begin{minipage}{0.49\linewidth}
\vspace{1pt}
\includegraphics[width=\textwidth]{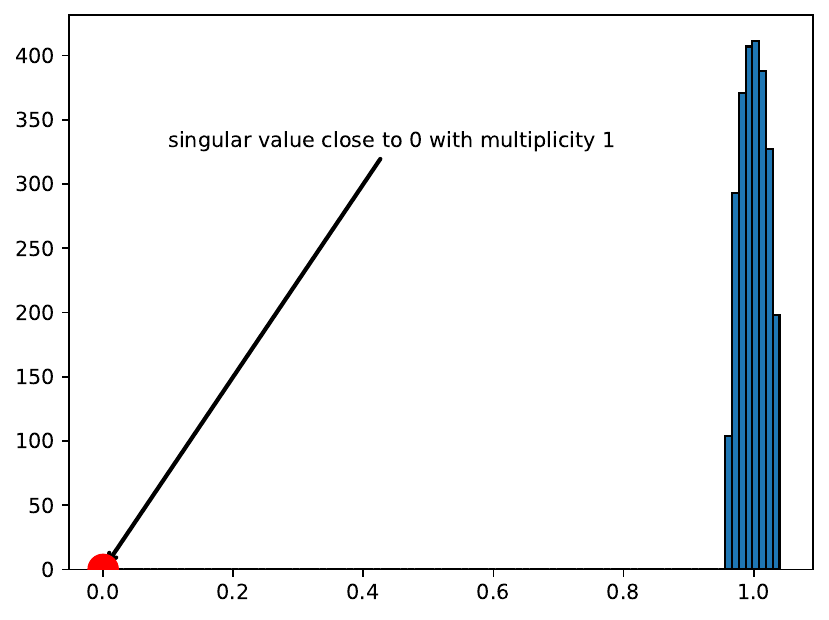}
\centerline{(b)}
\end{minipage}
\begin{minipage}{0.49\linewidth}
\vspace{1pt}
\includegraphics[width=\textwidth]{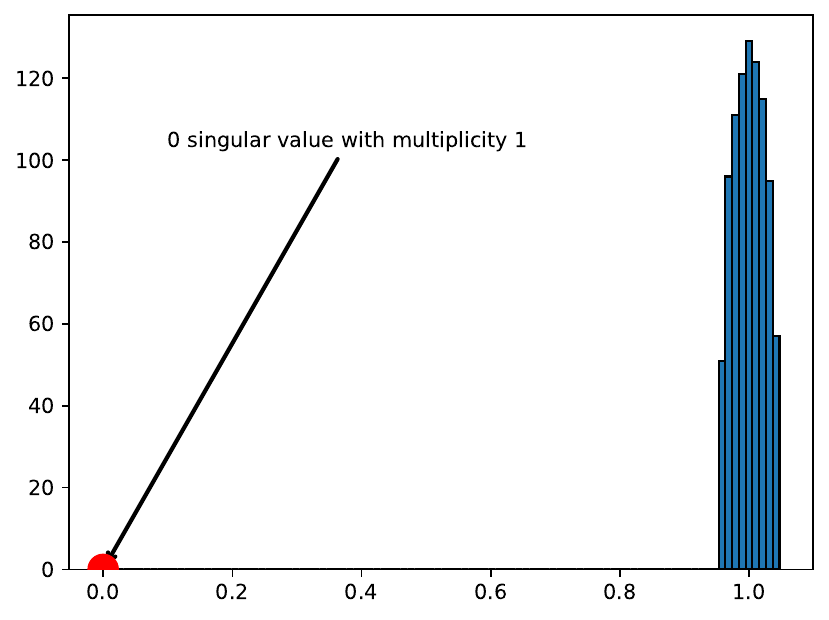}
\centerline{(c)}
\end{minipage}
\begin{minipage}{0.49\linewidth}
\vspace{1pt}
\includegraphics[width=\textwidth]{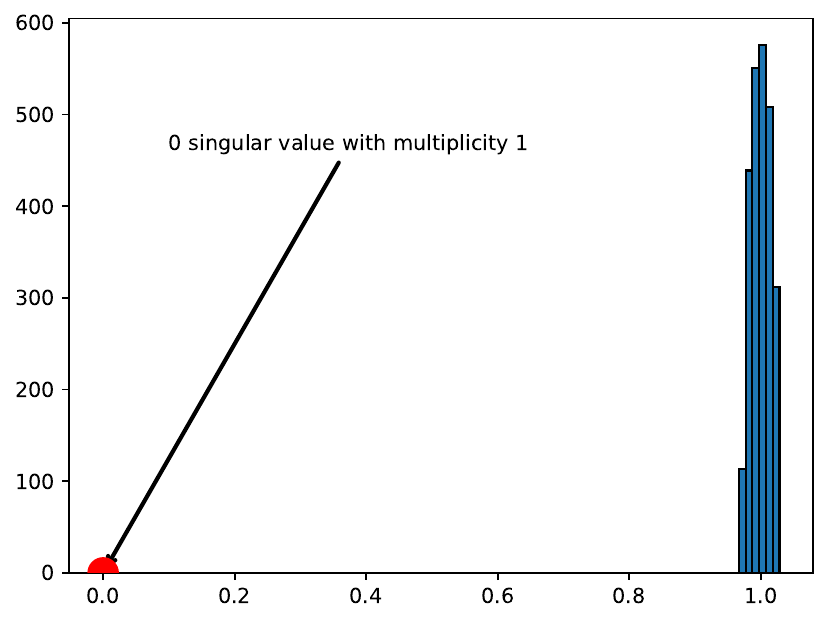}
\centerline{(d)}
\end{minipage}
\caption{Singular values behavior of $\tilde{B}+|\mathcal{S}|^{-1} ee^{\top}-I$ and $\mathfrak{P}-I$. (a) $|\mathcal{S}|=900$ for $\tilde{B}+|\mathcal{S}|^{-1} ee^{\top}-I$, (b) $|\mathcal{S}|=2500$ for $\tilde{B}+|\mathcal{S}|^{-1} ee^{\top}-I$, (c) $|\mathcal{S}|=900$ for $\mathfrak{P}-I$, (d) $|\mathcal{S}|=2500$ for $\mathfrak{P}-I$. }
\label{fig_eigenvalues_behavior_P}
\end{figure}

Next, we illustrate the singular value behavior of $W = \mathfrak{P} - I$ in Fig.\,\ref{fig_eigenvalues_behavior_P} and our estimates $\tilde{B} + |\mathcal{S}|^{-1} ee^{\top} - I$ for the uninformative prior case. It is shown that $W$ has exactly one singular value equal to 0, while the others are near 1. For $\tilde{B} + |\mathcal{S}|^{-1} ee^{\top} - I$, we observe that $|\mathcal{S}| - 1$ singular values are near 1, with fluctuations slightly larger than those of $W$, likely due to amplification of errors by the inequality.

\begin{figure}[htbp]
\centering
\begin{minipage}{0.36\linewidth}
\vspace{1pt}
\includegraphics[width=\textwidth]{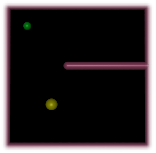}
\centerline{(a)}
\end{minipage}
\begin{minipage}{0.49\linewidth}
\vspace{1pt}
\includegraphics[width=\textwidth]{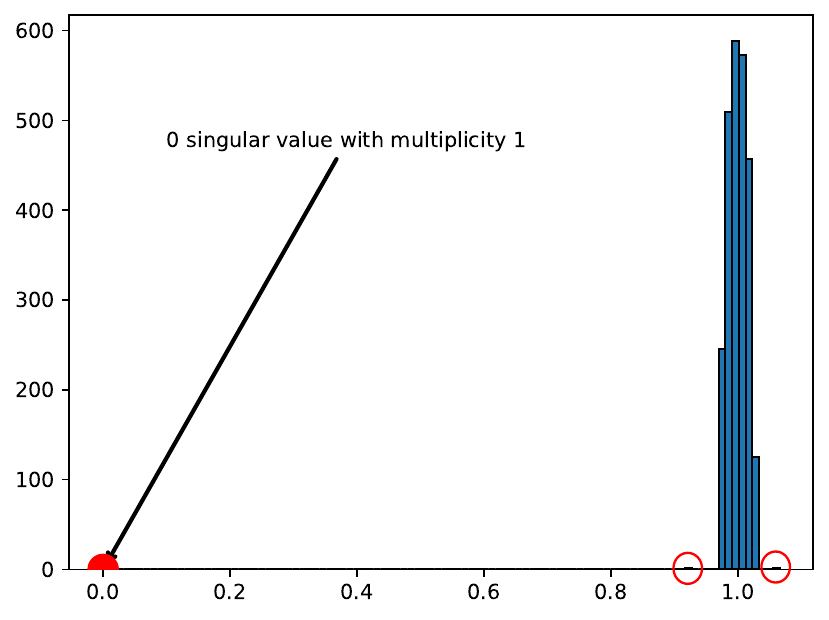}
\centerline{(b)}
\end{minipage}
\begin{minipage}{0.36\linewidth}
\vspace{1pt}
\includegraphics[width=\textwidth]{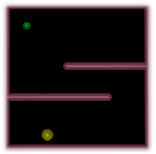}
\centerline{(c)}
\end{minipage}
\begin{minipage}{0.49\linewidth}
\vspace{1pt}
\includegraphics[width=\textwidth]{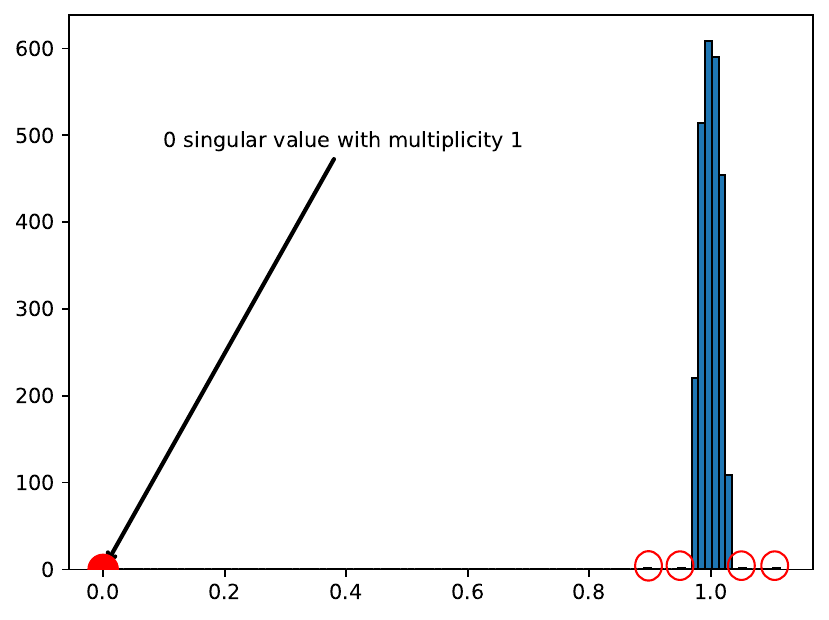}
\centerline{(d)}
\end{minipage}
\caption{Singular values behavior of $\mathfrak{P}-I$. (a) and (b): The situation of one barrier, (c) and (d): The situation of two barriers. }
\label{fig_informative_prior_barrier_experiments}
\end{figure}

Finally, we examine the singular value behavior of the informative prior case shown in Fig.\,\ref{fig_informative_prior_barrier_experiments}, focusing on the 2D maze environment \citep{brockman2016openai}. In this scenario, the prior transition probability of barrier position is 0, while other positions use uninformative prior. We observe similar singular value behavior as to the uninformative prior case. Notably, as the accumulated width of the barrier increases, the singular values move further away from 1, as indicated by the singular values in red empty circles. This phenomenon can be explained by the fact that when the accumulated width (dimension) is relatively large, we can no longer treat it as a finite rank case.

\subsection{Transfer Effect of PPO-AIRL + SAC\label{sec_transfer_effect_PPO-AIRL + SAC}}
In this subsection, we empirically assess the performance of PPO-AIRL + SAC. The experiment focuses on a 2D maze task, a quadrupedal ant agent. The 2D maze task navigates the agent (depicted in yellow) to achieve the goal (depicted in green), while the quadrupedal ant agent is trained to run forwards, which originates from Ant-v2 in OpenAI Gym \citep{brockman2016openai}. These tasks are widely used platforms in IL literature \citep{fu2018learning,qureshi2019adversarial,ni2020f}. Two cases are considered in the experiment. 

\begin{figure}[htbp]
\centering
\includegraphics[width=0.99\textwidth]{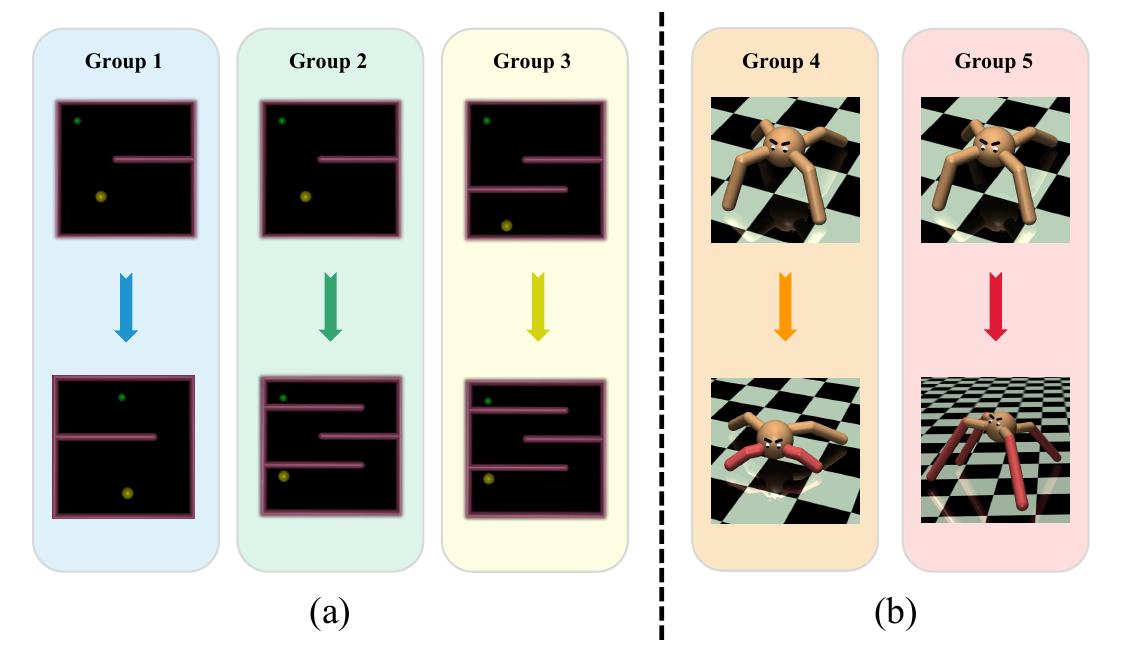}
\caption{Two cases of reward transfer scenarios. (a) Reward transfer scenarios under changes in the environment structure (Case 1). Group 1: A reward learned in PointMaze-Right is transferred to PointMaze-Left. Group 2: A reward learned in PointMaze-Right is transferred to PointMaze-Multi. Group 3: A reward learned in PointMaze-Double is transferred to PointMaze-Multi. (b) Reward transfer scenarios under changes in the agent dynamics (Case 2). Group 4: A reward learned in Ant is transferred to Ant-Disabled. Group 5: A reward learned in Ant is transferred to Ant-Lengthened. }
\label{fig_transfer_scenarios_environments}
\end{figure}

In Case 1, we validate the reward transfer performance when the alterations are made to the structure of the target environment. Consistent with \citep{fu2018learning}, in Group 1, PointMaze-Right is set to be the source environment and PointMaze-Left is taken to be the target environment. To further assess the robustness of the learned reward, we consider two more complicated environments (PointMaze-Double \& PointMaze-Multi) and add another two groups of experiments (Group 2 and Group 3). In Group 2, PointMaze-Right is set to be the source environment and PointMaze-Multi is taken to be the target environment. In Group 3, PointMaze-Double is set to be the source environment and PointMaze-Multi is taken to be the target environment. A visual display of these three groups of experiments is shown in Fig.\,\ref{fig_transfer_scenarios_environments}(a). 

In Case 2, we validate the reward transfer performance when the agent dynamics within the target environment are changed. Consistent with \citep{fu2018learning}, in Group 4, Ant is set to be the source environment and Ant-Disabled is taken to be the target environment. Analogous to Case 1, we add another group of experiments (Group 5). In Group 5, Ant-Lengthened is taken to be the target environment, where the legs are increased twice the length of those in Ant. A graphical depiction of both groups of experiments is presented in Fig.\,\ref{fig_transfer_scenarios_environments}(b). 

We evaluate PPO-AIRL + SAC against three control groups: PPO-AIRL + PPO, SAC-AIRL + SAC and SAC-AIRL + PPO. Additionally, for a comprehensive comparison, OPIRL \citep{hoshino2022opirl} and policy transfer by behavioral cloning (BC) \citep{pomerleau1991efficient,bain1995framework} (i.e., the policy imitated with BC in the source environment is directly applied to the target environment) are set as baselines to demonstrate the superiority of our results. We also incorporate two evaluation criteria: the random policy; and the training policy induced by the SAC algorithm with the ground truth reward in the target environment, which are referred to as ``Random policy'' and ``Oracle (SAC)'', respectively. 

\begin{figure}[htbp]
\centering
\begin{minipage}{0.32\linewidth}
\vspace{1pt}
\includegraphics[width=\textwidth]{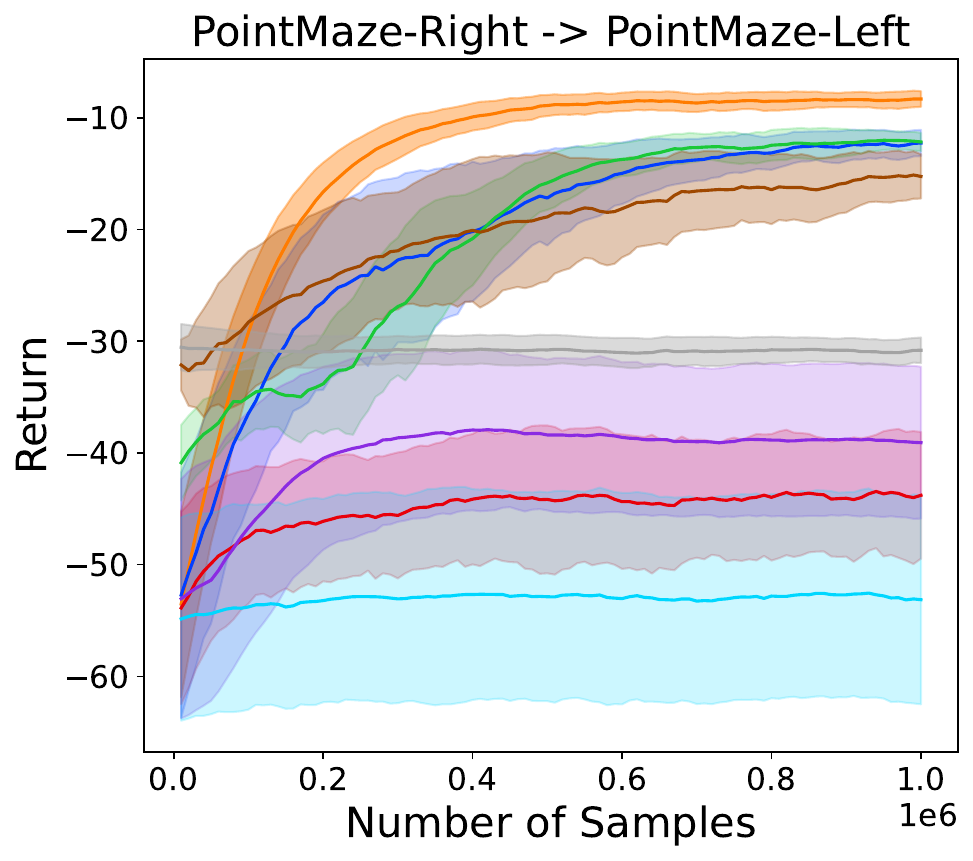}
\end{minipage}
\begin{minipage}{0.32\linewidth}
\vspace{1pt}
\includegraphics[width=\textwidth]{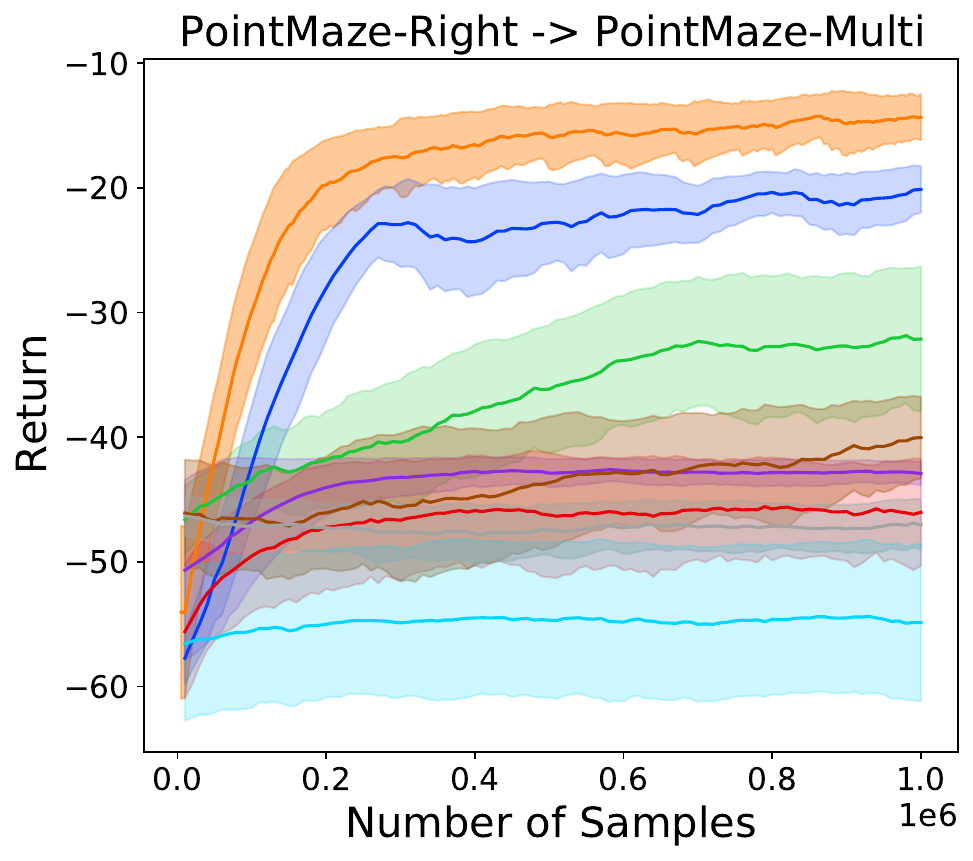}
\end{minipage}
\begin{minipage}{0.32\linewidth}
\vspace{1pt}
\includegraphics[width=\textwidth]{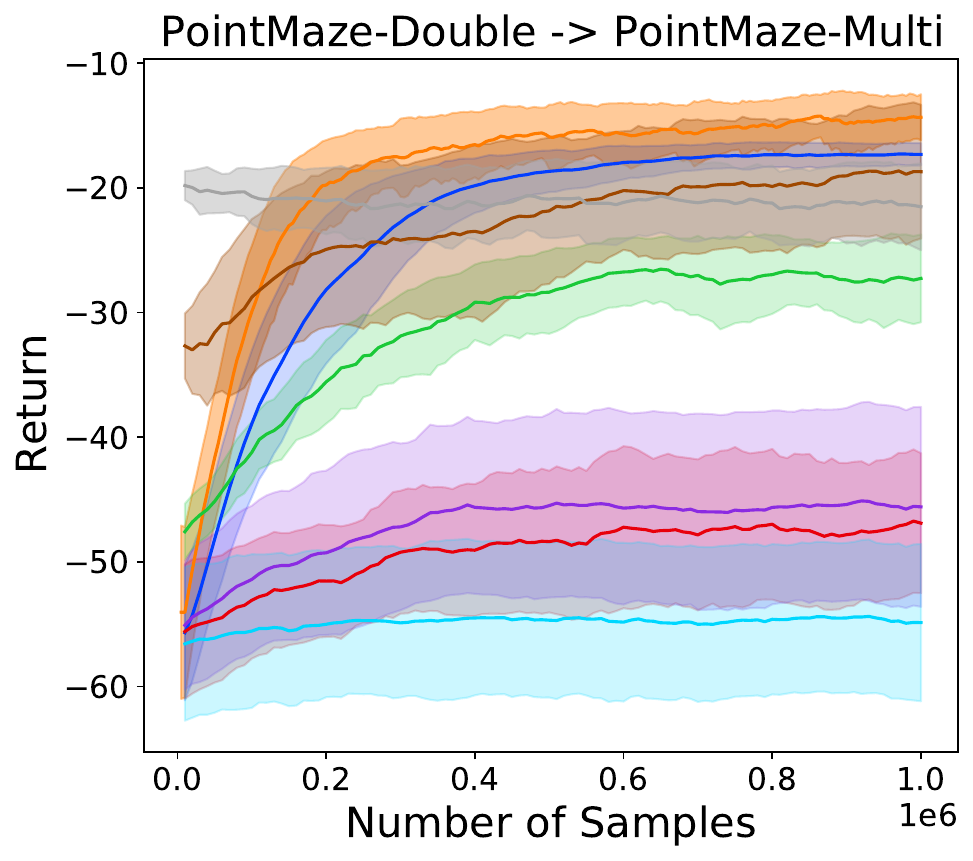}
\end{minipage}
\begin{minipage}{0.32\linewidth}
\vspace{1pt}
\includegraphics[width=\textwidth]{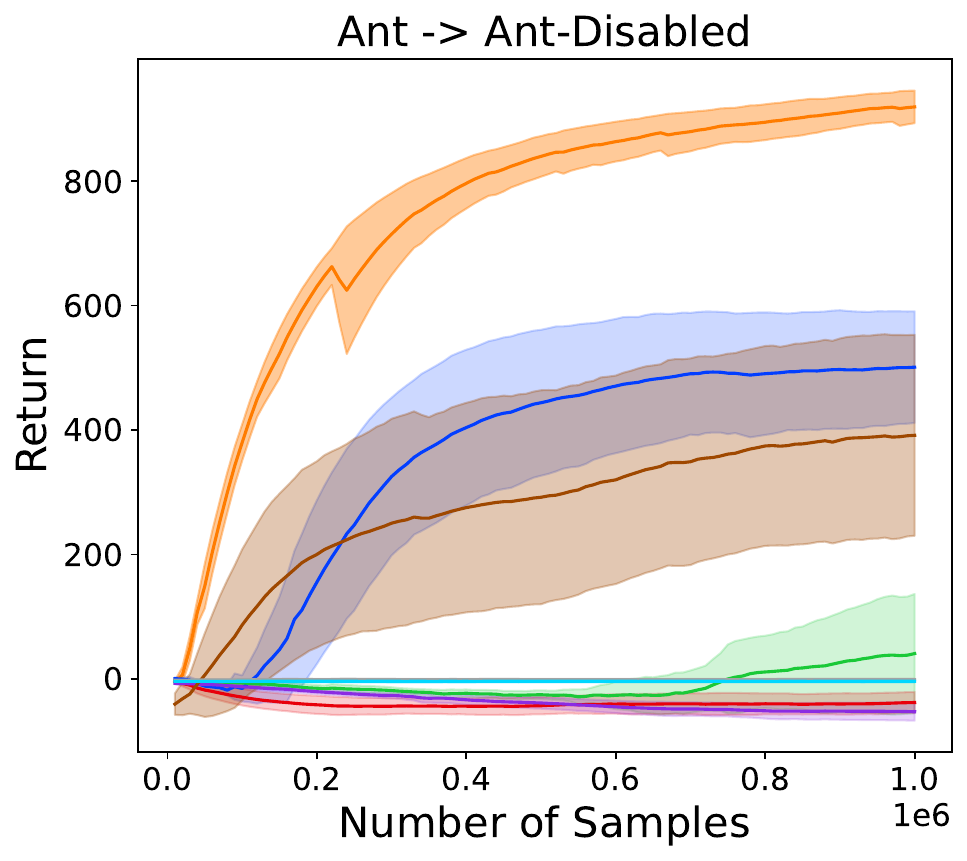}
\end{minipage}
\begin{minipage}{0.32\linewidth}
\vspace{1pt}
\includegraphics[width=\textwidth]{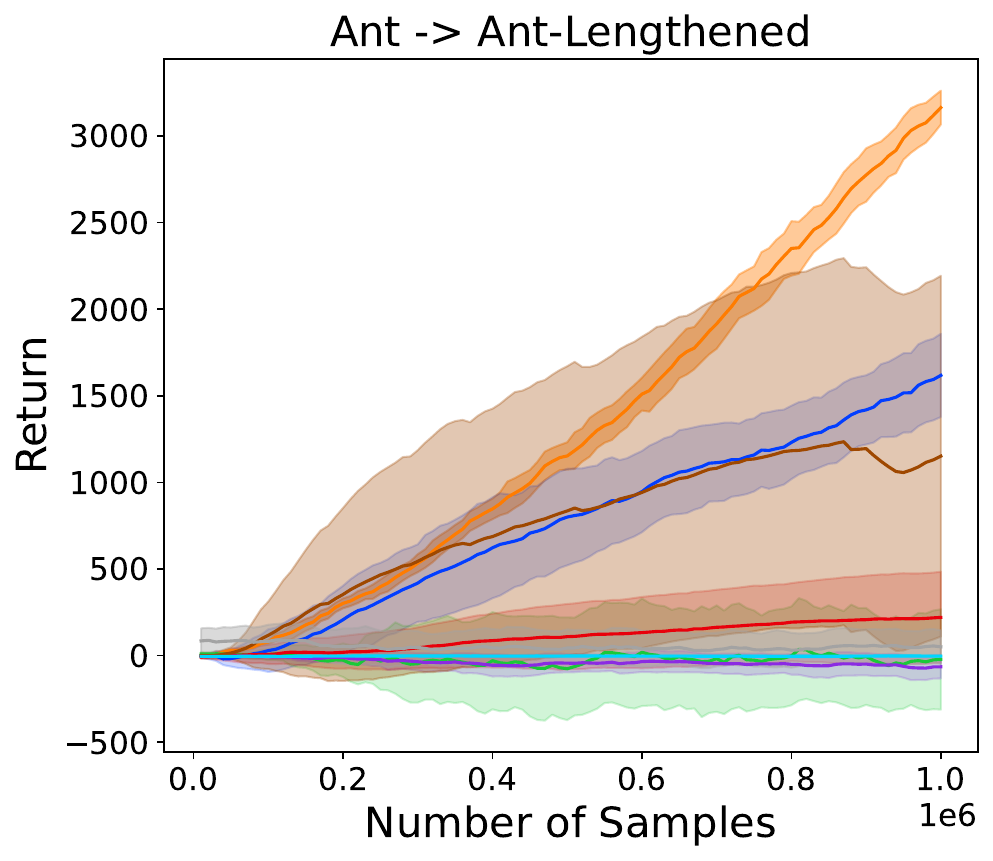}
\end{minipage}
\begin{minipage}{0.99\linewidth}
\vspace{1pt}
\centering
\includegraphics[width=\textwidth]{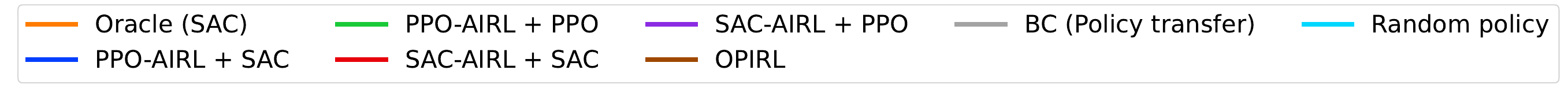}
\end{minipage}
\caption{Transfer performance with five seeds in the target environment. }
\label{transfer_experiment_pic}
\end{figure}

The performance of the eight algorithms is displayed in Fig.\,\ref{transfer_experiment_pic}. We discover that PPO-AIRL + SAC and PPO-AIRL + PPO surpass other methods in Group 1. Significantly, PPO-AIRL + SAC depicts a notably superior effect in all groups, closely matching ``Oracle (SAC)''. In contrast, SAC-AIRL based algorithms (SAC-AIRL + SAC and SAC-AIRL + PPO) exhibit poor performance. This observation aligns with the analysis that the reward extracted by SAC-AIRL cannot be exhaustively disentangled. Subsequently, the learned policy in the target environment benefits from the extensive exploration by SAC. In addition, it is worth noting that the performance of PPO-AIRL + SAC between Group 2 and Group 3 shows little difference (same target environment: PointMaze-Multi), which indicates the effectiveness of rewards disentangled by PPO-AIRL. 

\begin{remark}\label{remark_criticism_2}
Based on our experimental performance and theoretical analysis in Section \ref{sec_extractability_of_disentangled_rewards}, we argue that using SAC-AIRL in the source environment or PPO in the target environment--commonly considered baselines in prior studies \citep{arnob2020off,xu2022receding,hoshino2022opirl} - is unsuitable. Instead, we propose a more competitive framework: PPO-AIRL + SAC.
\end{remark}

\begin{figure}[htbp]
\centering
\begin{minipage}{0.32\linewidth}
\vspace{1pt}
\includegraphics[width=\textwidth]{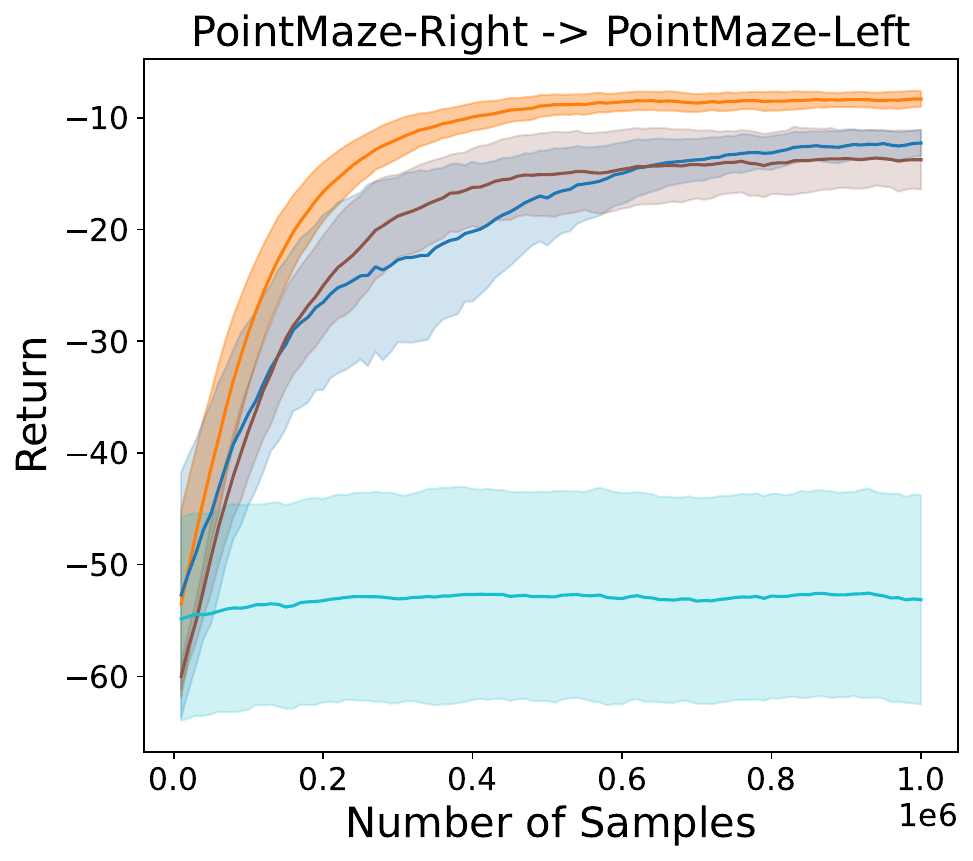}
\end{minipage}
\begin{minipage}{0.32\linewidth}
\vspace{1pt}
\includegraphics[width=\textwidth]{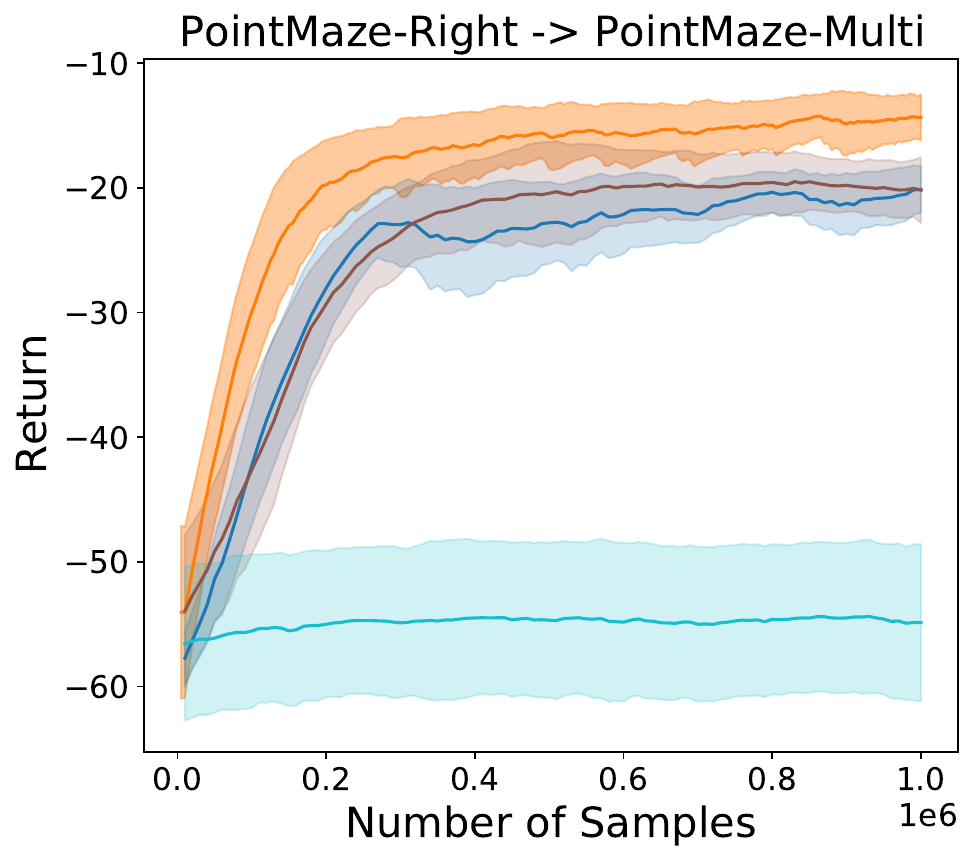}
\end{minipage}
\begin{minipage}{0.32\linewidth}
\vspace{1pt}
\includegraphics[width=\textwidth]{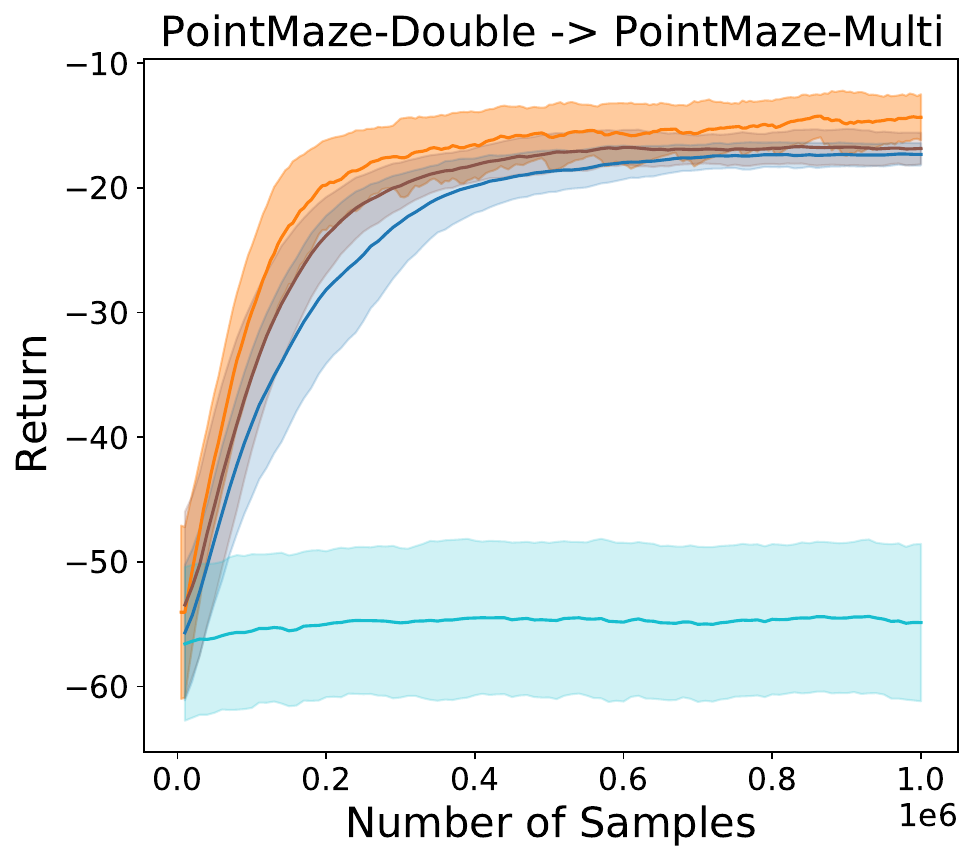}
\end{minipage}
\begin{minipage}{0.32\linewidth}
\vspace{1pt}
\includegraphics[width=\textwidth]{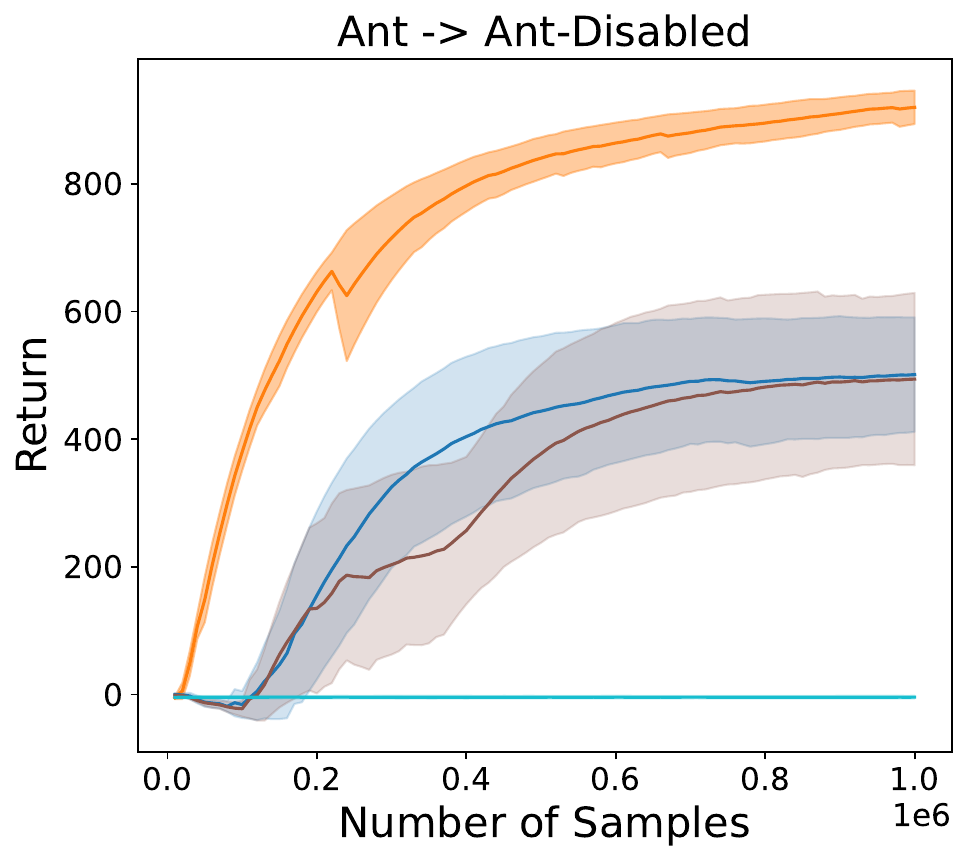}
\end{minipage}
\begin{minipage}{0.32\linewidth}
\vspace{1pt}
\includegraphics[width=\textwidth]{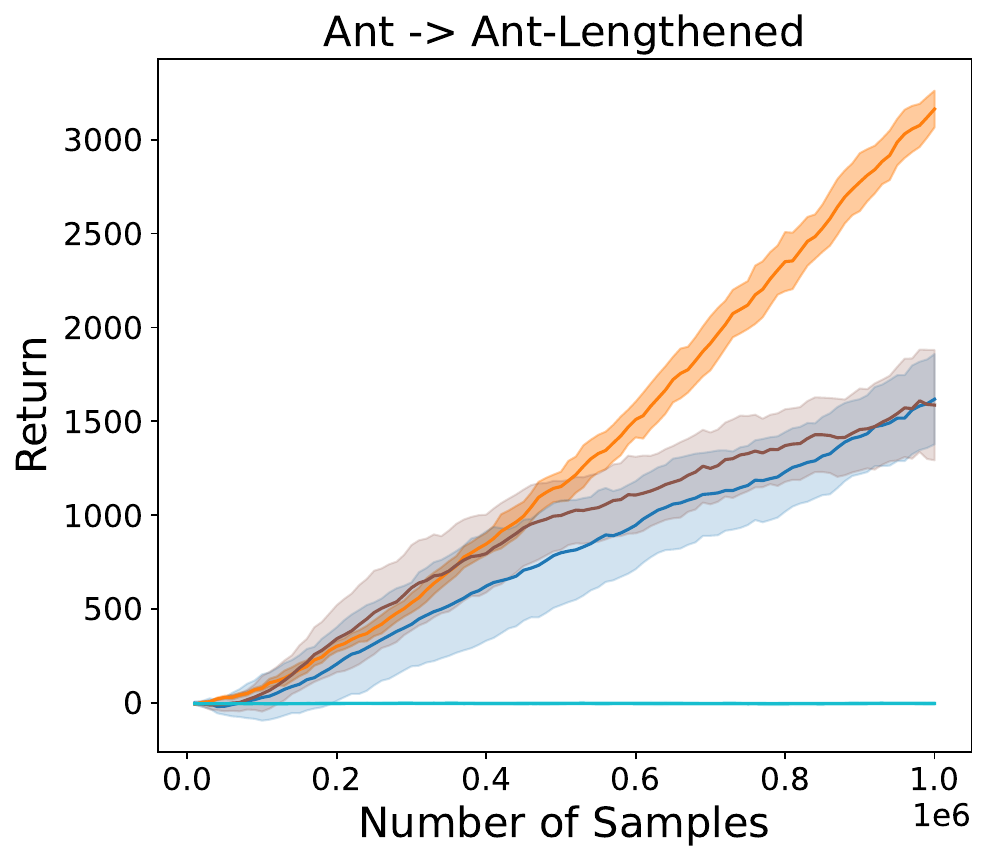}
\end{minipage}
\begin{minipage}{0.99\linewidth}
\vspace{1pt}
\centering
\includegraphics[width=\textwidth]{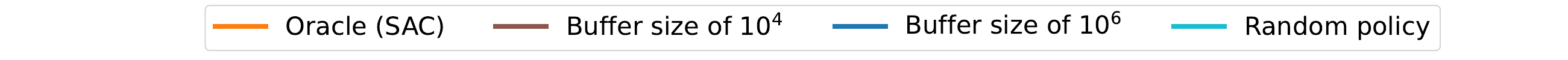}
\end{minipage}
\caption{Performance of different source expert buffer sizes with five seeds in the target environment. }
\label{buffer_size_ablation_pic}
\end{figure}

\noindent \textbf{Ablation study of different source expert buffer sizes.} Additionally, we conduct an ablation study on different source expert buffer sizes. The performance using expert demonstrations with buffer sizes of $10^{4}$ and $10^{6}$ in the source environment is shown in Fig.\,\ref{buffer_size_ablation_pic}, illustrating the robustness of our hybrid framework. 

\section{Conclusion\label{sec_conc}}
In this paper, we reanalyze the problem of reward transferability through the perspective of RMT, focusing on the implications of an unobservable transition matrix and extending our analysis to scenarios with limited prior knowledge. Our results reveal that, for both uninformative priors and limited prior information, the transferability condition \eqref{transferabilitycondition} is satisfied with high probability, enabling rewards to be identified within a constant factor. This insight bolsters the practical credibility of AIRL and shifts the explanation for inefficient transfer to a selection issue of its employed RL algorithm. We assess the performance of off-policy and on-policy RL algorithms in AIRL by quantifying training variance and propose the hybrid framework PPO-AIRL + SAC, which significantly improves reward transfer efficiency.

For future work, two key directions can be explored. First, the current theory constrains the non-zero singular values to be near 1, imposing a low-rank constraint on prior information. This limitation could be further explored and extended to accommodate higher-rank priors. Second, for another transfer paradigm involving $(s, a)$-shaped reward transfer (parallel to our state-only case) with multi-expert data, a more practical analysis could be conducted in scenarios where the state transition matrix is unobservable. We will investigate these two directions in future research.

\newpage
\appendix
\section{Proofs}
\textbf{Proof of Lemma \ref{Perturbation}:}
\begin{proof}
Lemma \ref{Perturbation} directly follows from the fact that ${\rm rank}(|\mathcal{S}|^{-1} ee^{\top})=1$ and rank inequality in \citep[Theorem A.43]{BSbook}, which states:
\begin{lemma}(Theorem A.43 in \citep{BSbook})
Let $A$ and $B$ be two $n \times n$ Hermitian matrices. Then,
\begin{align*}
\|F^A(x)-F^B(x)\|\leq \frac{1}{n}{\rm rank} (A-B). 
\end{align*}
\end{lemma}
\end{proof}

\textbf{Proof of Lemma \ref{maxeig}:}
\begin{proof}
Let $ M = B - \tilde{B} $ with its elements defined as
\begin{align*}
m_{ij} =
\begin{cases}
0 & \text{for } i \leq j, \\
b_{ij} - b_{ji} & \text{for } i > j.
\end{cases}
\end{align*}
Thus, we have $ \|B - \tilde{B}\|^2 = \|M\|^2 = \lambda_1(MM^\top) $. The elements of $ T := MM^\top $ are given by:
\begin{enumerate}
\item $ t_{11} = 0 $, and for $ i = 2, \ldots, |\mathcal{S}| $, we have $ t_{ii} = \sum_{j=1}^{i-1} m_{ij}^2 $,
\item $ t_{1j} = t_{j1} = 0 $, and for $ 2 \leq i < j \leq |\mathcal{S}| $, $ t_{ij} = t_{ji} = \sum_{k=1}^{i-1} m_{ik} m_{jk} $.
\end{enumerate}

Next, we want to determine the eigenvalue bounds for $ T $ using the Gerschgorin Circle Theorem. The Gerschgorin circles are centered at $ t_{ii} \prec |\mathcal{S}|^{-1} $, and we need to bound the radius $ R_i $. Without loss of generality, consider $ R_{|\mathcal{S}|} $, since the other radii behave similarly.

\begin{align*}
R_n = \sum_{i=2}^{|\mathcal{S}|} |t_{|\mathcal{S}|i}| = \sum_{i=2}^{|\mathcal{S}|} |t_{i|\mathcal{S}|}| = \sum_{i=2}^{|\mathcal{S}|} \Big|\sum_{k=1}^{i-1} m_{ik} m_{|\mathcal{S}|k}\Big|.
\end{align*}
Expanding $ m_{ik} $ and $ m_{|\mathcal{S}|k} $, we get:
\begin{align*}
R_n = \sum_{i=2}^{|\mathcal{S}|} \Big|\sum_{k=1}^{i-1} (b_{ik} - b_{ki})(b_{|\mathcal{S}|k} - b_{k|\mathcal{S}|})\Big|.
\end{align*}
Now, express $ b_{ij} $ in terms of $ x_{ij} $ and $ \bar{x}_i $, the average of $ x_{ij} $'s, to further bound the expression. We have:
\begin{align*}
R_n
&= \frac{1}{|\mathcal{S}|} \sum_{i=2}^{|\mathcal{S}|} \Big| \frac{1}{|\mathcal{S}|} \sum_{k=1}^{i-1} \Big(\bar{x}_i^{-1}(x_{ik} - \bar{x}_i) - \bar{x}_k^{-1}(x_{ki} - \bar{x}_k)\Big)\Big(\bar{x}_{|\mathcal{S}|}^{-1}(x_{|\mathcal{S}|k} - \bar{x}_{|\mathcal{S}|}) - \bar{x}_k^{-1}(x_{k|\mathcal{S}|} - \bar{x}_k)\Big) \Big|\\
&=\frac{1}{|\mathcal{S}|}\sum_{i=2}^{|\mathcal{S}|}\Big|\frac{1}{|\mathcal{S}|}\sum_{k=1}^{i-1}\Big(\bar{x}_i^{-1}(x_{ik}-x_{ki})+x_{ki}(\bar{x}_i^{-1}-\bar{x}_k^{-1})\Big)\Big(\bar{x}_{|\mathcal{S}|}^{-1}(x_{|\mathcal{S}|k}-x_{k|\mathcal{S}|})+x_{k|\mathcal{S}|}(\bar{x}_{|\mathcal{S}|}^{-1}-\bar{x}_k^{-1})\Big)\Big|. 
\end{align*}
Given the uniform approximations $ |\bar{x}_i^{-1} - 1| \prec |\mathcal{S}|^{-1/2} $ and $ |\bar{x}_i^{-1} - \bar{x}_k^{-1}| \prec |\mathcal{S}|^{-1/2} $ from \eqref{maxxi}, we now examine the following terms:
\begin{align*}
\frac{1}{|\mathcal{S}|} \sum_{k=1}^{i-1}(x_{ik} - x_{ki}) \quad \text{and} \quad \frac{1}{|\mathcal{S}|} \sum_{k=1}^{i-1} (x_{ik} - x_{ki})(x_{|\mathcal{S}|k} - x_{k|\mathcal{S}|}).
\end{align*}
By the independence of the $ x_{ik} $'s, $x_{|\mathcal{S}|k}$'s, and the fact that
\begin{align*}
\mathbb{E}(x_{ik} - x_{ki}) = \mathbb{E}((x_{ik} - x_{ki})(x_{|\mathcal{S}|k} - x_{k|\mathcal{S}|})) = 0,
\end{align*}
we can apply the same method used in \eqref{profD-I} to obtain:
\begin{align*}
\frac{1}{|\mathcal{S}|} \sum_{k=1}^{i-1}(x_{ik} - x_{ki}) \prec |\mathcal{S}|^{-1/2}, \quad \frac{1}{|\mathcal{S}|} \sum_{k=1}^{i-1}(x_{ik} - x_{ki})(x_{|\mathcal{S}|k} - x_{k|\mathcal{S}|}) \prec |\mathcal{S}|^{-1/2}.
\end{align*}
Thus, we conclude that:
\begin{align*}
R_n \prec |\mathcal{S}|^{-1/2}.
\end{align*}
Consequently, the Gerschgorin circles are centered at $ t_{ii} \prec |\mathcal{S}|^{-1} $, and the radii satisfy $ R_i \prec |\mathcal{S}|^{-1/2} $. This gives 
\begin{align*}
\|B - \tilde{B}\| = \sqrt{\lambda_1(T)} \prec |\mathcal{S}|^{-1/4}, 
\end{align*}
completing the proof of Lemma \ref{maxeig}.
\end{proof}

\textbf{Proof of Lemma \ref{bound}:} 
\begin{proof}
We rewrite $QQ^{\top}$ as 
\begin{align}\label{Sn} 
|\mathcal{S}|^{-1}X\left(I - |\mathcal{S}|^{-1}ee^{\top}\right)X^{\top}, 
\end{align} 
where $X = (x_{ij}) \in \mathbb{R}^{|\mathcal{S}| \times |\mathcal{S}|}$. Under the deterministic shift $x_{ij} \to x_{ij} - f_i$, we find that $QQ^{\top}$ remains invariant:
\begin{align*}
&\left(X-\text{Diag}(f_1,\ldots,f_{|\mathcal{S}|})ee^{\top}\right)\left(I-|\mathcal{S}|^{-1}ee^{\top}\right)\left(X-\text{Diag}(f_1,\ldots,f_{|\mathcal{S}|})ee^{\top}\right)^{\top},\\
=&X(I-|\mathcal{S}|^{-1}ee^{\top})X^{\top}-X\left(I-|\mathcal{S}|^{-1}ee^{\top}\right)\left(\text{Diag}(f_1,\ldots,f_{|\mathcal{S}|})ee^{\top}\right)^{\top}\\
&-\text{Diag}(f_1,\ldots,f_{|\mathcal{S}|})ee^{\top}\left(I-|\mathcal{S}|^{-1}ee^{\top}\right)X^{\top}\\
&+\text{Diag}(f_1,\ldots,f_{|\mathcal{S}|})ee^{\top}\left(I-|\mathcal{S}|^{-1}ee^{\top}\right)\left(\text{Diag}(f_1,\ldots,f_{|\mathcal{S}|})ee^{\top}\right)^{\top}\\
=&X(I-|\mathcal{S}|^{-1}ee^{\top})X^{\top}.
\end{align*}
The last equality holds since $\text{Diag}(f_1,\ldots,f_{|\mathcal{S}|})ee^{\top}\left(I-|\mathcal{S}|^{-1}ee^{\top}\right)=0$.
Therefore, by setting $f_i = 1$, this is equivalent to studying the following paradigm:
\begin{align}
\mathbb Ex_{ij}=0,~~\mathbb Ex_{ij}^2=1,~~\text{and}~~\mathbb E|x_{ij}|^q\leq C,
\label{ass1}
\end{align}
for any sufficiently large constant $q$ and some constant $C$. 

The ESD of $QQ^{\top}$ is given by
\begin{align*}
F_{|\mathcal{S}|}(x)=\frac{1}{|\mathcal{S}|}\sum_{j=1}^{|\mathcal{S}|} {\mathbb I}_{\{\lambda_{j}(QQ^{\top}) \leq x\}},\quad x\in\mathbb{R},
\end{align*}
and the Stieltjes transform of $F_{|\mathcal{S}|}$ is 
\begin{align*}
m_{|\mathcal{S}|}(z)=\int \frac{1}{x-z}{\rm d}F_{|\mathcal{S}|}(x),
\end{align*}
where $z=E+i\eta \in \mathbb{C}^{+}$.

Based on i.i.d. $x_{ij}$'s and the paradigm \eqref{ass1}, we have the following proposition describing the global location of eigenvalues:

\begin{proposition}\label{global location}(Global location)
Suppose that $QQ^{\top}$ satisfies Assumption \ref{ass1}. Then, as $|\mathcal{S}|\to \infty$, $F_{|\mathcal{S}|}$ almost surely converges to a probability distribution $F_{\delta_1}$, whose Stieltjes transform $m_{\delta_1}=m_{\delta_1}(z)$ is determined by
\begin{align}\label{th11}
m_{\delta_1}=\frac{-z+\sqrt{(z-2)^2-4}}{2z}. 
\end{align}
\end{proposition}
This is precisely the Stieltjes transform of the Mar{\v{c}}enko-Pastur (M-P) law \citep{marvcenko1967distribution}, with a density function given by 
\begin{align}\label{density}
F^{'}_{\delta_1}(x)=\frac{1}{2\pi x}\sqrt{(4-x)(x-0)}.
\end{align}
We find the right edge of the support of $F^{'}_{\delta_1}(x)$, $\gamma_{+}=4$.

We denote $\gamma_1\geq\cdots\geq \gamma_{|\mathcal{S}|}$ as the ordered $|\mathcal{S}|$-quantiles of $F_{|\mathcal{S}|}$, i.e., $\gamma_{j}$ is the smallest real number such that
\begin{eqnarray}\label{local law1}
\int_{-\infty}^{\gamma_{j}}{\rm d}F_{\delta_1}(x)=\frac{|\mathcal{S}|-j+1}{|\mathcal{S}|},~j=1,\ldots,|\mathcal{S}|.
\end{eqnarray}
Then, for small positive $ c, \epsilon$ and sufficiently large $C_+ > 4$, we define the domain
\begin{eqnarray}\label{domin}
D(c,\epsilon):=\{z=E+i\eta\in \mathbb C^{+}:4- c\leq E\leq C_+,~|\mathcal{S}|^{-1+\epsilon}\leq \eta\leq 1\}.
\end{eqnarray}

The main tool in the study of the local location of eigenvalues is the Green function or resolvent, defined as
\begin{align}\label{Green Function}
G(z)=(QQ^{\top}-zI)^{-1},
\end{align}
with elements $G_{ij}$ for $i,j=1,\ldots,|\mathcal{S}|$. The Stieltjes transform of $F_{|\mathcal{S}|}$ is linked with Green function \eqref{Green Function} by
\begin{align*}
m_{|\mathcal{S}|}(z)=\frac{1}{|\mathcal{S}|}\sum_{j=1}^{|\mathcal{S}|} \frac{1}{\lambda(QQ^{\top})-z}=\text{Tr} G(z).
\end{align*}

We now assert that the following local law holds.
\begin{proposition}\label{local law}(Local law)
Under Assumption \ref{ass1}, we have that for any sufficiently small $\epsilon>0$,

\begin{enumerate}
\item (Entrywise local law): 
\begin{align*}
G_{ij}(z)=m_{\delta_1}(z)\delta_{ij}+O_{\prec}(\Psi(z))
\end{align*}
holds uniformly over $i,j=1,\ldots |\mathcal{S}|$ and $D(c,\epsilon)$, where $\delta_{ij}$ denotes the Kronecker delta, i.e. $\delta_{ij}=1$ if $i=j$, and $\delta_{ij}=0$ if $i\neq j$. Here, we defined the deterministic error parameter $\Psi(z):=\sqrt{\frac{{\rm {\rm Im}}m_{\delta_1}(z)}{|\mathcal{S}|\eta}}+\frac{1}{|\mathcal{S}|\eta}$. 
\item (Average local law): 
\begin{align*}
m_{|\mathcal{S}|}(z)= m_{\delta_1}(z)+O_{\prec}(\frac{1}{|\mathcal{S}|\eta})
\end{align*}
holds uniformly on $D(c,\epsilon)$.
\item (Eigenvalue Rigidity): 
\begin{align*}
|\lambda_i(QQ^{\top})-\gamma_i|\prec |\mathcal{S}|^{-\frac{2}{3}}\left(i\wedge(|\mathcal{S}|+1-i)\right)^{-1/3}.
\end{align*}
uniformly on $i=1,\ldots,|\mathcal{S}|$.
\end{enumerate}
\end{proposition}

Proposition \ref{global location} is actually a direct consequence of Theorem 1 in \citep{marvcenko1967distribution}, while Proposition \ref{local law} is a trivial extension of Theorem 3.6 in \citep{knowles2017anisotropic} specifically for square sample matrices (i.e., $M=N=|\mathcal{S}|$) and an identity population (i.e., $T=I$). Notably, there is no need to establish a local law between $m_{|\mathcal{S}|}(z)$ and the non-asymptotic version of $m_{\delta_1}(z)$, since the dimension-scale ratio and the ESD of the population covariance matrix satisfy
\begin{align*}
\frac{M}{N}=\frac{|\mathcal{S}|}{|\mathcal{S}|}\equiv 1\rightarrow 1,~~H_{|\mathcal{S}|}(x)\equiv \delta_1\rightarrow \delta_1,
\end{align*}
and the convergence rate is faster than that of the eigenvalues.

We derive from Proposition \ref{global location} that $\gamma_1=4$, and in combination with the Eigenvalue Rigidity in Proposition \ref{local law}, it suffices to prove Lemma \ref{bound}.
\end{proof}

\bibliographystyle{apalike}
\bibliography{ref}

\end{document}